\documentclass{article}

\PassOptionsToPackage{numbers}{natbib}

\usepackage[final]{neurips_2021}

\usepackage[utf8]{inputenc} 
\usepackage[T1]{fontenc}    
\usepackage{hyperref}       
\usepackage{url}            
\usepackage{booktabs}       
\usepackage{amsfonts}       
\usepackage{nicefrac}       
\usepackage{microtype}      
\usepackage{xcolor}         

\usepackage{thmtools}
\usepackage{thm-restate}

\usepackage{graphicx}

\usepackage{times}
\usepackage{amsmath}
\usepackage{bm}
\usepackage{amsthm}
\usepackage{multirow}
\usepackage[super]{nth}
\usepackage{amssymb}%
\usepackage{mathrsfs}%
\usepackage{pifont}
\usepackage{array}
\usepackage{wrapfig}
\usepackage{adjustbox}

\usepackage{kotex}
\usepackage{algorithm}
\usepackage{algorithmic}

\usepackage[normalem]{ulem}
\usepackage[labelformat=simple]{subcaption}

\usepackage{caption}
\usepackage{wrapfig}
\newtheorem{theorem}{Theorem}[section]

\newcolumntype{a}{>{\columncolor{yellow}}c}

\usepackage{amsmath,amssymb}
\usepackage{graphicx}
\usepackage{color}
\usepackage{bm}
\usepackage{algorithm}
\usepackage{algorithmic}
\usepackage{wrapfig}
\usepackage{hyperref}
\usepackage{url}

\newcommand{\bbox}{\hfill $\Box$}

\newcommand{\T}{ {\mathrm{\scriptscriptstyle T}} }

\newcommand{\benr}{\begin{eqnarray}}
\newcommand{\eenr}{\end{eqnarray}}
\newcommand{\benrr}{\begin{eqnarray*}}
\newcommand{\eenrr}{\end{eqnarray*}}
\newcommand{\ben}{\begin{equation}}
\newcommand{\een}{\end{equation}}
\newcommand{\benn}{\begin{equation*}}
\newcommand{\eenn}{\end{equation*}}

\newcommand{\pa}{\partial}

\newcommand{\vep}{\varepsilon}

\newcommand{\cD}{\mathcal D}
\newcommand{\cF}{\mathcal F}
\newcommand{\cG}{\mathcal G}
\newcommand{\cH}{\mathcal H}
\newcommand{\cL}{\mathcal L}

\newcommand{\cY}{\mathcal Y}

\newcommand{\bbI}{\mathbb I}

\def\rva{{\mathbf{a}}}

\def\rvx{{\mathbf{x}}}
\def\rvy{{\mathbf{y}}}

\newcommand{\E}{\mathbb{E}}

\newcommand{\R}{\mathbb R}

\DeclareMathOperator*{\argmax}{arg\,max}

\def\sP{{\mathbb{P}}}

\title{MixACM: Mixup-Based Robustness Transfer via Distillation of Activated Channel Maps}

\author{%
  Muhammad Awais \textsuperscript{\rm 1, 2}\thanks{Equal Contribution.}\, \thanks{This work was carried out at Huawei Noah’s Ark Lab. The webpage for the project is available at: {\color{blue}\href{https://awaisrauf.github.io/MixACM}{awaisrauf.github.io/MixACM}}}, 
  Fengwei Zhou \textsuperscript{\rm 1}\footnotemark[1], 
  Chuanlong Xie \textsuperscript{\rm 1}\footnotemark[1], 
  Jiawei Li \textsuperscript{\rm 1}, \\
  \textbf{Sung-Ho Bae \textsuperscript{\rm 2}\thanks{Corresponding Author.}, 
  Zhenguo Li \textsuperscript{\rm 1}}
  \\
  ~\\
  \textsuperscript{\rm 1} Huawei Noah's Ark Lab \\
  \textsuperscript{\rm 2} Department of Computer Science, Kyung-Hee University, South Korea \\
  \texttt{awais@khu.ac.kr}, \texttt{\{zhoufengwei, xie.chuanlong, li.jiawei\}@huawei.com}, \\
  \texttt{shbae@khu.ac.kr}, \texttt{li.zhenguo@huawei.com} \\
}

\begin{document}

\maketitle

\begin{abstract}
Deep neural networks are susceptible to adversarially crafted, small and imperceptible changes in the natural inputs. The most effective defense mechanism against these examples is adversarial training which constructs adversarial examples during training by iterative maximization of loss. The model is then trained to minimize the loss on these constructed examples. This min-max optimization requires more data, larger capacity models, and additional computing resources. It also degrades the standard generalization performance of a model. Can we achieve robustness more efficiently? In this work, we explore this question from the perspective of knowledge transfer. First, we theoretically show the transferability of robustness from an adversarially trained teacher model to a student model with the help of mixup augmentation. Second, we propose a novel robustness transfer method called Mixup-Based Activated Channel Maps (MixACM) Transfer. MixACM transfers robustness from a robust teacher to a student by matching activated channel maps generated without expensive adversarial perturbations. Finally, extensive experiments on multiple datasets and different learning scenarios show our method can transfer robustness while also improving generalization on natural images.      
\end{abstract}

\section{Introduction}

Deep learning models have achieved impressive performance on a wide variety of challenging tasks such as image recognition, natural language generation, game playing, etc. 
However, these models are susceptible to even small changes in the input space. It is possible to craft tiny changes in the input such that a model classifies unaltered input correctly but classifies the same input incorrectly after small and visually imperceptible perturbations~\cite{szegedy2013intriguing, goodfellow2014explaining}. These altered inputs are known as adversarial examples, and this vulnerability of the state-of-the-art models has raised serious concerns~\cite{kurakin2016adversarial, papernot2017practical, brown2017adversarial, ma2021understanding}. 

Many defense mechanisms have been proposed to train deep models to be robust against such adversarial perturbations. These techniques include defensive distillation \cite{papernot2016distillation}, gradient regularization \cite{gu2014towards, papernot2017practical, ross2018improving}, model compression \cite{das2018compression, liu2018security}, activation pruning \cite{dhillon2018stochastic, rakin2018defend}, adversarial training \cite{madry2017towards}, etc. Among them, Adversarial Training (AT) is one general strategy that is the most effective \cite{athalye2018obfuscated}. In general, adversarial training is a kind of data augmentation technique that trains a model on examples augmented with adversarial changes \cite{madry2017towards}. The adversarial training consists of an inner, iterative maximization loop to augment natural examples with adversarial perturbations, and an outer minimization loop similar to normal training. Many different methods have been introduced to improve robustness \cite{uesato2019labels, carmon2019unlabeled, najafi2019robustness, zhai2019adversarially, wong2020fast, shafahi2019adversarial, zhang2019theoretically, wang2019improving, bai2021improving,  wu2020adversarial, kim2020adversarial, lin2020dual, pang2020boosting}, but all of them are fundamentally based on the principle of training on adversarially augmented examples. 

Adversarial training is more challenging compared with normal training. Better robust generalization requires larger capacity models \cite{madry2017towards, nakkiran2019adversarial, wu2020does}. Even over-parameterized models that can easily fit data for normal training \cite{zhang2016understanding} may have insufficient capacity to fit adversarially augmented data \cite{zhang2020geometry}. The sample complexity of adversarial training can be significantly higher than normal training, and it requires more labeled \cite{schmidt2018adversarially} or unlabeled data \cite{uesato2019labels, carmon2019unlabeled, najafi2019robustness, zhai2019adversarially}. The inner maximization for the generation of adversarial perturbations for adversarial training is significantly more expensive computationally as it requires iterative gradient steps with respect to the inputs (e.g., adversarial training takes $\sim$7x more time compared with normal training). Adversarial training also degrades the performance of a model on natural examples significantly \cite{nakkiran2019adversarial, zhang2019theoretically}.

Can we attain robustness more efficiently, i.e., with less data, small capacity models, without extra back-propagation steps, and at no significant sacrifice of clean accuracy? For normal training, knowledge transfer is one possible paradigm for the efficient training of deep neural networks. In normal knowledge transfer, a pre-trained teacher model is leveraged to efficiently train a student model on the same or similar dataset \cite{srinivas2018knowledge}. However, knowledge transfer methods are designed to transfer features related to normal generalization which may be at odds with robustness \cite{tsipras2018robustness, ilyas2019adversarial}. 
A recent line of work has explored transferring or distilling robustness from pre-trained models \cite{hendrycks2019pretraining, shafahi2019adversarially, goldblum2019adversarially, chan2020thinks}. Although these techniques are effective, they also require extra gradient computation and may not work in some cases. 

In this work, we argue that a better approach for robustness transfer is to distill intermediate features of a robust teacher, generated on mixup examples. Our proposed approach does not require any additional gradient computation and works well with smaller models and fewer data samples. It can also effectively transfer robustness across models and datasets.

We begin with the theoretical analysis and show that adversarial loss of a student model can be bounded with two terms: natural loss and distance between the student and teacher model on mixup examples. To minimize the distance between robust teacher and student, we proposed a new distillation method. Our proposed distillation method is based on channel-wise activation analysis of robustness by \citet{bai2021improving} and other studies showing that adversarially trained models learn fundamentally different features~\cite{ilyas2019adversarial, tsipras2018robustness, zhang2019interpreting}. 

Channels of a convolutional neural network learn different disentangled representations which, when combined, describe specific semantic concepts~\cite{bau2020units}. 
Samples of different classes activate different channels for an intermediate layer. However, adversarial examples make these channels activate more uniformly thereby destroying class-related information. Adversarial training solves this issue by forcing a similar channel-activation pattern for normal and adversarial examples~\cite{bai2021improving}. Our proposed distillation method generates activated channel maps from a robust teacher which has already learned robust channel-activation patterns. The student, then, is forced to match these activated channel maps. 

We have conducted extensive experiments to show the effectiveness of our method using various datasets and under different learning settings. We start by exhibiting the ability of our method to transfer robustness without requiring adversarial examples. We then show that our method can distill robustness from large pre-trained models to smaller models and it can transfer robustness across datasets. We also show that our method is capable to robustify a model even with a small number of examples. Concisely, our contributions are as follows:
\begin{enumerate}

    \item We show that adversarial loss of a student model can be bounded with the distance between robust teacher and student model on mixup examples. 

    \item We propose a new method to transfer robustness from the intermediate features of robust pre-trained models. 

    \item We have demonstrated effectiveness of our method with experiments on CIFAR-10, CIFAR-100, and ImageNet datasets. We showed that MixACM can achieve adversarial robustness comparable to the state-of-the-art methods without generating adversarial examples. Our method also improves clean accuracy significantly.
\end{enumerate}
\vfill

\section{Related Work}

\textbf{Adversarial Training and Robust Features.}
Adversarial attacks are considered to be a security concern~\cite{biggio2013evasion, szegedy2013intriguing, goodfellow2014explaining, carlini2017towards}, and many methods have been proposed to defend against such attacks~\cite{gu2014towards, goodfellow2014explaining, papernot2016distillation, guo2017countering, papernot2017practical,tramer2017ensemble, madry2017towards, buckman2018thermometer, liao2018defense, ross2018improving, das2018compression, liu2018security, dhillon2018stochastic, rakin2018defend}. Adversarial training~\cite{madry2017towards} is the most effective defense~\cite{athalye2018obfuscated} among these methods. The fundamental principle of adversarial training is to generate adversarial perturbation during training by doing iterative back-propagation w.r.t. input. The model is then trained on these perturbed examples. Building upon this principle, different aspects of adversarial training have extensively been studied, e.g., effect of large unlabeled data~\cite{uesato2019labels, carmon2019unlabeled, zhai2019adversarially}, theoretical aspects to improve the trade-off between accuracy and robustness~\cite{zhang2019theoretically}, improving parts of adversarial training for better robustness~\cite{wang2019improving, wu2020adversarial, pang2020boosting, bai2021improving, gowal2020uncovering}, making adversarial training fast~\cite{najafi2019robustness, shafahi2019adversarial, wong2020fast, andriushchenko2020understanding}, pruning with adversarial training \cite{ye2019adversarial, sehwag2020hydra}, adversarial training with unlabeled data~\cite{kim2020adversarial}, etc. Our work is different from these methods as our main concern is robustness transfer without generating adversarial perturbations. Another line of work studied the effects of adversarial training and showed that adversarial training learns `fundamentally different'~\cite{nakkiran2019adversarial} features compared with normal training~\cite{ilyas2019adversarial, xu2019interpreting, zhu2021towards}. Our method is motivated by these studies and is designed to leverage these robust features.

\textbf{Knowledge and Robustness Transfer.}
Knowledge transfer is training a student model more efficiently on a dataset with the help of a teacher model trained on a similar or related dataset~\cite{srinivas2018knowledge}. Many different settings have been explored to achieve this objective by using soft labels produced by teacher~\cite{bucilu2006model, hinton2015distilling}, knowledge learned by intermediate layers of teacher also called feature distillation~\cite{romero2014fitnets, zagoruyko2016paying, huang2017like, yim2017gift, tung2019similarity, koratana2019lit, shen2019meal, heo2019knowledge, heo2019comprehensive, tian2019contrastive}, or matching input gradients~\cite{zagoruyko2016paying, srinivas2018knowledge}. However, the main objective of these knowledge transfer methods is improving the generalization on natural, unperturbed images. Our work is fundamentally different from these works as we want to transfer robustness. 

Our work is motivated by~\cite{hendrycks2019pretraining, shafahi2019adversarially, goldblum2019adversarially, chan2020thinks, awais2021adversarial} showing transferability of robustness. \citet{hendrycks2019pretraining} showed that robust features learned on large datasets like ImageNet can improve adversarial training. \citet{shafahi2019adversarial} showed that robust pre-trained models can act as feature extractors in transfer learning. These two methods are limited in application as pre-trained models are used as backbone models. In~\cite{goldblum2019adversarially}, the authors showed that the robustness can be distilled more efficiently from a large pre-trained teacher model to a smaller student model by using the teacher-produced class scores on adversarial examples. Compared with \cite{goldblum2019adversarially}, our method does not require any adversarial training. In~\cite{chan2020thinks}, the distillation is performed by matching the gradient of the teacher and student, requiring fine-tuning teacher on target dataset and back-propagation to get gradients of the teacher and student w.r.t. inputs and training of a discriminator. Compared with~\cite{chan2020thinks}, our proposed method does not require any back-propagation steps in addition to those carried out in normal training.

\textbf{Activation and Robustness.}
Many works have explored the role of activations in the robustness from different angles, such as modifying output of intermediate layers to improve adversarial training \cite{xiao2019resisting,dhillon2018stochastic,madaan2019adversarial,mustafa2020deeply,xie2019feature}, using different activation functions \cite{rakin2018defend,wang2018adversarial,xie2020smooth,gowal2020uncovering,pang2020bag,bai2021improving, awais2020towards} or interpretation of what is learned by adversarial training with activations \cite{xu2019interpreting, bai2021improving}. However, we leveraged activations to transfer robustness. Our distillation method is inspired by the channel-wise analysis of robustness proposed by \citet{bai2021improving}.

\textbf{Mixup Augmentation and Robustness.}
Mixup augmentation was introduced to improve the generalization however, it also improved performance against one-step adversarial attacks~\cite{zhang2017mixup}. This motivated many mixup-based adversarial defense methods, e.g., ~\citet{pang2019mixup, lee2020adversarial, archambault2019mixup, bunk2021adversarially, chen2021guided}. These defense methods primarily improve adversarial training by mixup. Our work, however, is different as we explore the role of the mixup in robustness transfer from both theoretical and algorithmic perspectives. Our theoretical work is based on~\cite{zhang2020does} that showed a connection between mixup and robustness theoretically. \citet{goldblum2019adversarially} also briefly empirically explored the effect of some augmentation methods on robustness transfer.

\section{Setup}
\label{sec:setup}

We consider task of mapping input $x \in \mathcal{X} \subseteq \mathbb{R}^d$ to label $y \in \mathcal{Y}=\{1,2,\ldots,K\}$. 
Given the training data $\cD = \{(x_1, y_1), ..., (x_n, y_n)\}$, the goal is to learn a classifier $f: \mathcal{X} \to \R^k$ from a hypothesis space $\cF.$ Let $L\big(f(x), y\big)$ be the loss function that measures that how poorly the classifier $f(x)$ predicts the label $y.$
Suppose $f$ can be decomposed into $g\circ h$, where $h \in \cH$ is a feature extractor and $g \in \cG$ stands for the classifier on top of feature extractor. 

\textbf{Adversarial Augmentation}: Adversarial training augments natural examples with adversarial perturbations which are attained by maximizing the loss, i.e., $\tilde{x} = x + \delta$ where $\delta = \argmax_{\|\delta'\|_p \leq \epsilon} L\big( f(x+\delta'), y \big).$ Here $\|\cdot\|_p$ stands for $L_p$ norm. Throughout this paper, we take $p=\infty.$ 
We denote the adversarial loss \cite{madry2017towards} as: 
\benrr
\tilde \cL(f, \cD)  = \frac{1}{n} \sum_{i=1}^n \max_{\|\delta\|_\infty \leq \epsilon} L(f(x_i+\delta), y_i),
\eenrr
where $\epsilon$ is the perturbation budget that governs the adversarial robustness of the model.

\textbf{Mixup Augmentation}: Mixup constructs virtual examples by linearly combining two examples \cite{pang2019mixup}: 
$\tilde{x}_{ij}(\lambda) = \lambda x_i + (1-\lambda) x_j, $ and $\tilde{y}_{ij}(\lambda) = \lambda y_i + (1-\lambda) y_j, $ where 
$\lambda \in [0, 1]$ follows the distribution $P_\lambda.$ The mixup loss is 
\[
\cL_{mix}(f, \cD) =
\frac{1}{n^2} \sum_{i=1}^n \sum_{j=1}^n \mathbb{E}_{\lambda \sim  P_{\lambda}} [ L( f(\tilde{x}_{ij}(\lambda)), \tilde{y}_{ij}(\lambda))].
\]

\section{Analysis: How to Transfer Robustness?}

In this section, we provide an analysis on how to transfer robustness with knowledge distillation and mixup augmentation. We first explore this problem from a theoretical perspective and show that student's adversarial loss can be decomposed into the sum of the teacher's adversarial loss and the distillation loss. Furthermore, these two loss functions can be approximated by mixup-based augmentation. We then discuss how we can design a distillation loss that distills robustness from the intermediate features of a robust model.

\subsection{Theoretical Analysis}
\label{sec:theoretical_analysis}

\textbf{Robustness Transfer.} We give a generalization bound for the robust test error via distillation.
Different to the classical bound, we compress the hypothetical space $\cF$ into a smaller space 
derived by the teacher feature extractor.
\citet{hsu2021generalization} points out that the distillation helps to derive a fine-grained analysis for the prediction risk and avoid getting a vacuous generalization bound. 
Therefore, we take the following theorem as a starting point to derive robustness transfer.
We consider the loss function $L(f(x), y) = 1 - \phi_\gamma\big( f(x), y\big)$, where $\phi_\gamma$ is the softmax function with temperature $\gamma>0$:
\benrr
\phi_\gamma\big( f(x), y\big) = \frac{\exp(f(x)_y / \gamma)}{\sum_{k=1}^K \exp(f(x)_k / \gamma)}.
\eenrr
\begin{theorem}
Let the temperature $\gamma>0$ be given.
Then, with probability $1-\delta$, for any $f \in \cF$, 
\benr\label{main}
\E\big[\max_{\|\delta\|_\infty \leq \epsilon} \bbI\{\hat y(x+\delta)\neq y\}\big]
\leq 2\tilde \cL(f, \cD) + \frac{8}{\gamma}\tilde \cL_{dis}(f, \cD)  + 4 \mathfrak{R}_S(\Psi_T) + \frac{16}{n}+ 6\sqrt{\frac{2/\delta}{2n}},
\eenr
where $\mathfrak{R}_S$ is the Rademacher complexity, $\Psi_{T}= \{\max_{\|\delta\|_\infty<\epsilon} L(g\circ h^T(x+\delta),y), \,\, g\in\cG\}$ and
\benrr
\tilde \cL_{dis}(f,\cD) = \frac{1}{n} \inf_{g \in \cG } \sum_{i=1}^n \max_{\|\delta\|_\infty<\epsilon} \big| f(x_i+\delta) - g \circ h^T(x_i+\delta) \big|.
\eenrr
\end{theorem}

On the right hand side of (\ref{main}), two terms depend on the classifier $f$, i.e., the adversarial loss $\tilde \cL(f, \cD)$ and the distillation loss $\tilde \cL_{dis}(f, \cD).$ 
Thus, we can minimize $2 \tilde \cL (f, \cD) +8 \tilde \cL_{dis}(f, \cD)/\gamma$ to find a classifier with small robust test error. 
Note that the loss function $L$ is bound above by the cross-entropy loss. So $\tilde \cL(f, \cD)$ can be replaced with a common-used adversarial loss based on cross-entropy. 

\textbf{Adversarial Robustness, Transfer and Mixup.}
However, solving the maximization problem for $\delta$ is time-consuming, which is a severe problem in adversarial training especially for large models.
For binary classification task with logistic loss function, \citet{zhang2020does} proved that $\tilde \cL(f, \cD)$ can be bounded above by $\cL_{mix}(f,\cD).$ 
In addition, the distillation loss that measures the distance between the student model and the teacher model also considers the worst-case $\delta$. 
In our algorithm, we use the KL divergence or Euclidean distance and mixup sample to design the distillation loss.
Inspired by the Theorem~3.3 in \cite{zhang2020does}, we present the following example to illustrate the rationality behind our design.

\begin{theorem}
Suppose that $\cY=\{0,1\}$ and $L$ is the logistic loss. Let $f$ be a fully connected neural network with ReLU activation function or max-pooling. 
Suppose a learnt teacher model $f^T$ with the same network structure is given.
Denote $\cD_{dis} = \{\big(x_i, f^T(x_i)\big), i=1,\ldots,n\}.$
Under certain assumptions, we have for any positive scalar $\alpha$,
\benrr
\tilde \cL(f, \cD) + \alpha \tilde \cL(f, \cD_{dis}) \leq  \cL_{mix}(f, \cD) + \alpha \cL_{mix}(f, \cD_{dis}).
\eenrr
\end{theorem}

This result shows that both the adversarial loss and the distillation loss can be bounded above by their mixup loss. Please see Appendix.B for more details and complete proof. Here the distillation loss is formulated by the prediction accuracy for  pseudo-labels generated by a learnt teacher model, which is also considered by \cite{carmon2019unlabeled}.

\begin{figure}[t]
    \centering
    \includegraphics[width=1\linewidth]{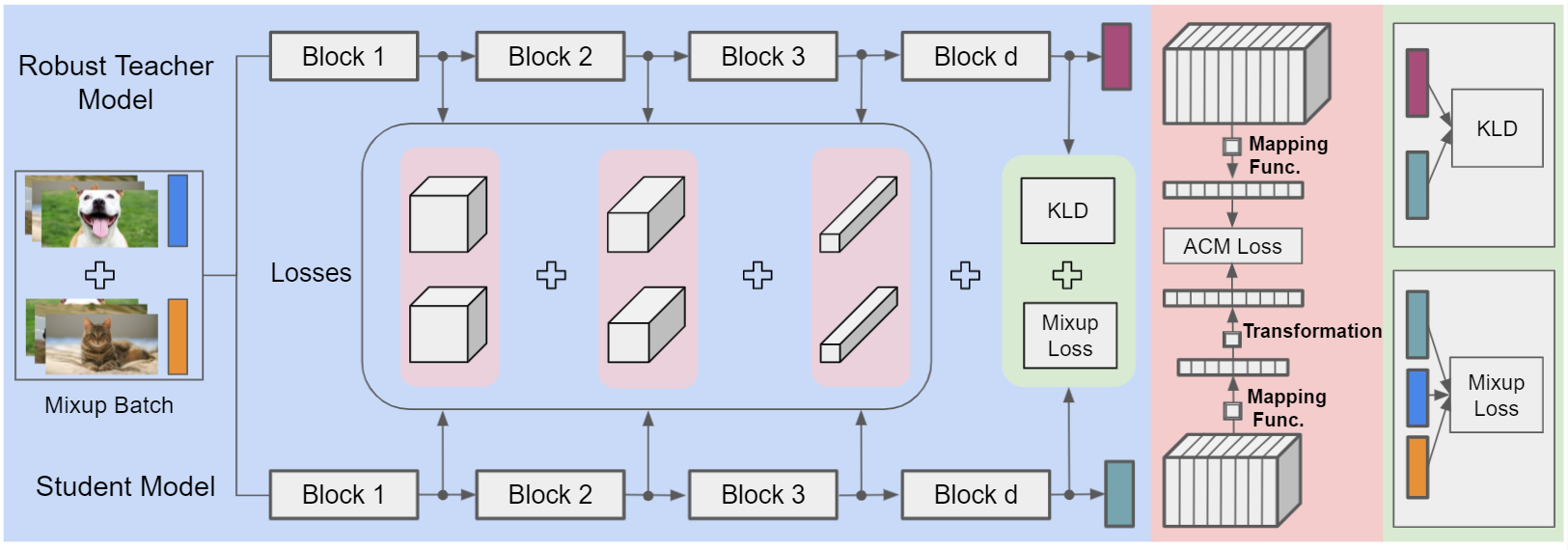}
    \caption{\textbf{Overview of our approach}. The mixup augmented examples are passed through a robust teacher and a student to get intermediate features. These features are then passed through a mapping function to get activated channel maps. The student mimics the activated channel maps of robust teacher by minimizing ACM loss. The student model also minimizes the standard KLD and Mixup loss.}
    \label{fig:overview}
    \vspace{-10pt}
\end{figure}

\subsection{Considerations for Algorithmic Design}
In order to transfer robustness, our theoretical analysis suggests minimizing the distance between teacher and student on mixup examples. However, as observed by \citet{goldblum2019adversarially} and our experiments in Section~\ref{sec:experiment}, minimization of KL-Divergence between the outputs of teacher and student models is not enough. How can we reduce the gap between the teacher and student more effectively? Based on previous work~\cite{engstrom2019adversarial, xu2019interpreting, bai2021improving}, we hypothesize that the gap between latent space of robust teacher and student is large and it requires distillation from intermediate features.

To reduce the distance between the latent spaces of robust teacher and student, we get inspiration from recent channel-based analysis of robustness \cite{bai2021improving} and works on the interpretation of what is learned by robust models \cite{nakkiran2019adversarial, ilyas2019adversarial, xu2019interpreting}. Specifically, we consider convolution channels of CNNs to be fundamental learning blocks: each channel learns class-specific patterns and the output of the model is a combination of these basic patterns. As noted by \cite{ilyas2019adversarial}, these basic blocks learn both robust and non-robust features with natural training. Adversarial examples amplify non-robust features while suppressing the robust features, e.g., adversarial examples frequently activate channels that are rarely activated by natural examples \cite{bai2021improving}. Adversarial training works by suppressing non-robust channels and making the activation frequency of natural and adversarial examples similar.

Motivated by this analysis, we propose Activated Channel Map (ACM) transfer loss. Our method extracts the activated channel maps of the robust model's intermediate features on mixup examples. The student is then forced to mimic these maps. Specifically, we consider the maximum value of each channel as a proxy for the extent of how activated a channel is: a channel is less activated if it has a smaller maximum value and more activated if it has a high maximum value. We then make the student match the normalized activation map of the teacher. And, further analysis of our method is presented in the supplementary material. We present the core of our method in the next section.

\section{Methodology}
\begin{wrapfigure}{r}{0.25\textwidth}
    \includegraphics[width=0.25\textwidth]{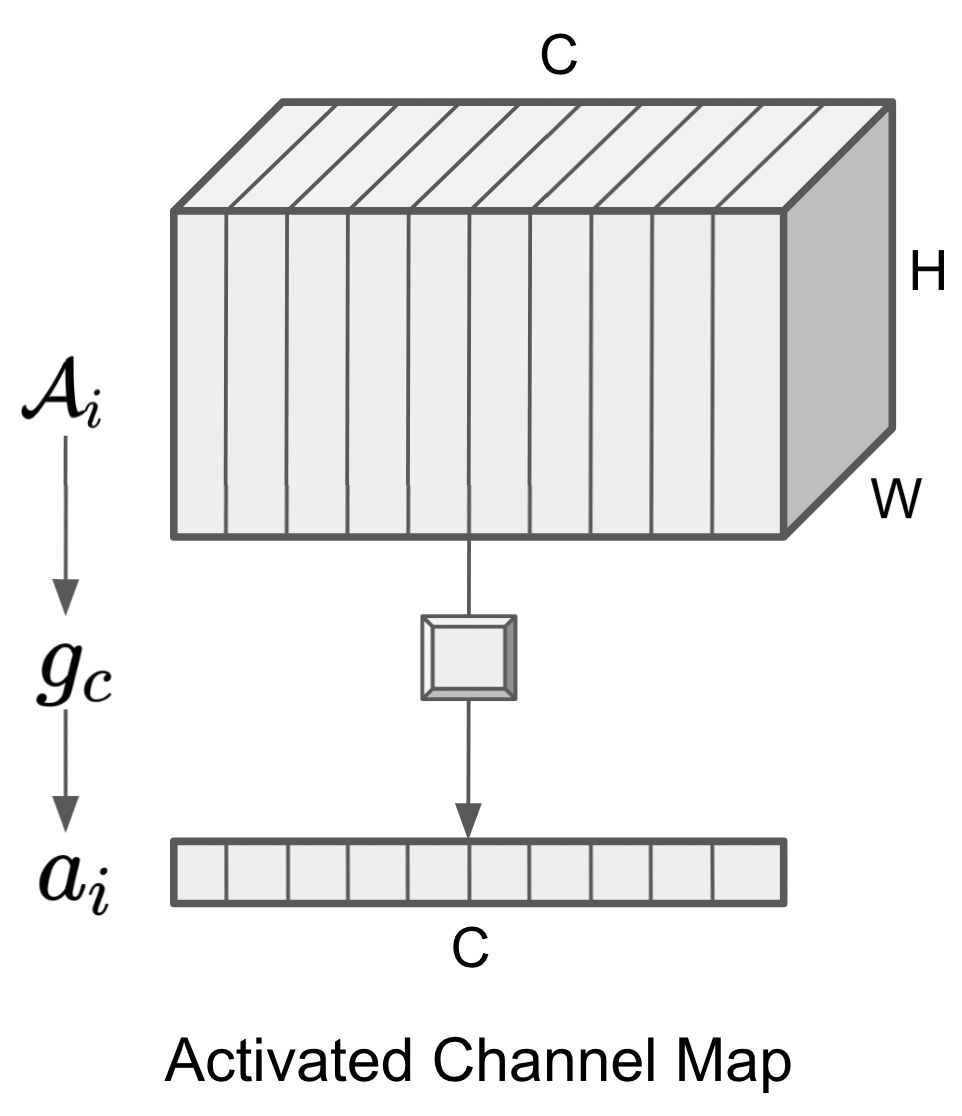}
    \caption{Mapping function $g_c$ is applied channel-wise to get the Activated Channel Maps (ACM).}
    \label{fig:phi_c}
    \vspace{-15pt}
\end{wrapfigure} 
\subsection{Activated Channels Maps (ACM) Transfer Loss}
In this section, we introduce our Mixup-Based Robustness Transfer method. As discussed above, the method aims to transfer robustness from a robust teacher model $f^T$ to a student model $f^S$ via matching activated channel maps of a robust teacher on mixup augmented examples. An overview of our method is given in Figure \ref{fig:overview}.

The purpose of robust teacher is to supervise the student's focus on robust features. This is achieved by matching activated channel maps of teacher and student. The activated channel maps of a model is extracted as follows: given $i$-th block activations $\mathcal{A}_i$ (output of $i$-th layer), we apply a function $g_c: \mathbb{R}^{B\times C \times H \times W} \to \mathbb{R}^{B\times C \times 1 \times 1}$ to get the activated channel map ${a}_i = g_c(\mathcal{A}_i)$, as illustrated in Figure \ref{fig:phi_c}.

The activated channel map of student and teacher may differ in scale. The contribution of each channel in the final output, on the other hand, depends only on the relative value compared with other channels. Therefore, we first normalize this map. To force the student to mimic the teacher's activated channel maps, we use $\ell_2$ distance between the normalized activated channel maps of teacher and student. For $i$-th layer, we can define ACM loss as: 
\[
\mathcal{L}_{acm_i}(\mathcal{A}^T_i, \mathcal{A}^S_i) = \| \dfrac{{a}^T_i}{\|{a}^T_i\|_2} -  \dfrac{{a}^S_i}{\|{a}^S_i\|_2}\|_2^2.
\]

We minimize this distance at the end of each block of the models. A block is a set of convolution layers grouped like residual blocks in ResNets \cite{he2016deep}. For a student and teacher model with $d$ such blocks, the total ACM loss becomes as follows: 
$\mathcal{L}_{acm}(x; f^T, f^S) = \sum_{i=0}^d \mathcal{L}_{acm_i}(f^T(x)_i, f^S(x)_i), $
where $f(x)_i$ represents output of the model at $i$-th block. 

\textbf{Mapping Function ($g_c$)}: The aim of activated channel map function ($g_c$) is to extract a map of least to most activated channels. For this purpose, we used maximum value of each channel: $g_c(x) = \max_i x_i \forall{i} \in \{0, ..., c_l\} $, where $c_l$ is number of channels in $l$-th layer. 

\subsection{Transferring Robustness with Different Feature Sizes}
 Our method extracts a scalar value corresponding to each channel in a layer. Until now, we assume that the teacher and student have the same number of channels. For this case, we only need the $g_c$ mapping function. To accomplish knowledge transfer across models with a different number of channels, we need to apply transformation either on a teacher or student's activated channel maps. We experimented with two transformations: adaptive pooling and affine transformation. 

\textbf{Adaptive Pooling}:
 Pooling is used in most CNN architectures to reduce the size of feature vectors and gain invariance to small transformations of the input. The pooling layer works by dividing the input into patches and taking an average or maximum value over these patches. By changing the size of these patches, pooling can also be used for up-sampling. Pooling operation can be described as follows: $y[p] = \max_{q\in \Omega(p)} x[q]$, where $\Omega$ represents a patch or neighbourhood of $x[p]$. We use PyTorch implementation of 1d adaptive maximum pooling to either down or up-sample teacher or student output. 

\textbf{Affine Transformation}: 
We can also use the learning function to convert the input size to output size. For this, we use a fully connected layer. Based on our ablation studies described in supplementary material, we use adaptive pooling as a default in our method.

\subsection{KL Divergence Loss with Soft Labels}
Soft labels have shown to be useful for adversarial robustness by \citet{pang2020bag}. Therefore, for distillation on the same dataset (e.g., where teacher and student have the same classification layer), we minimize the KL divergence between the logits provided by teacher and student following \cite{hinton2015distilling, goldblum2019adversarially}. The KLD loss becomes as follows:
$
\mathcal{L}_{kld}(x; f^T, f^S) =  \gamma ^2 KL(y^T(\gamma), y^S(\gamma)),
$
where $\gamma$ is the temperature parameter, and $y(\gamma)$ is output logits of a model normalized by temperature parameter as used by \citet{hinton2015distilling}. We omit this loss term when transferring from one dataset to another. 

\subsection{Mixup Fueled Distillation} 
Our theoretical analysis suggest minimizing distance between teacher and student with mixup examples. To this end, we use mixup to linearly combine inputs and outputs and minimize loss on these inputs. The final training objective for the student is as follows:
\[\min_{f^S}  \mathbb{E}_{(x,y)\sim \mathcal{D}} \mathbb{E}_{\lambda \sim P_{\lambda}} \big[ (1-\alpha_{kld})\mathcal{L}_{mix} + \alpha_{kld} \mathcal{L}_{kld}+ \alpha_{acm} \mathcal{L}_{acm} \big] ,\]

where $\alpha_{kld}$ and $\alpha_{acm}$ are hyperparameters for corresponding loss terms. 

\section{Experiments}
\label{sec:experiment}
In this section, we empirically evaluate our method for robustness transfer. We primarily show the effectiveness of our method under three settings: 1) Robustness Transfer in Section \ref{sec:robustness_transfer} to demonstrate that our method can transfer robustness from a pre-trained teacher to student without generating adversarial examples; 2) Robustness transfer from larger to smaller models in Section \ref{sec:robustness_compression} to show its effectiveness under distillation settings; and 3) robustness transfer under transfer learning settings, e.g., from one dataset to another dataset and robustness with smaller dataset sizes in Section \ref{sec:robustness_transfer_learning}.  

\begin{table}[t]
\centering
\caption{\textbf{Detailed Comparison}. CIFAR-10 test accuracy and robustness comparison under two different $\ell_\infty$ attacks of magnitude $\epsilon=8/255$. Our method outperforms robustness distillation methods in both accuracy and robustness. It outperforms various existing state-of-the-art robust models in clean accuracy while retaining comparable robustness. All reported values for our method are mean $\pm$ std of five repetitions. B.P. in the table stands for back propagation.}

\begin{adjustbox}{width=1\textwidth}
\begin{tabular}{lcccccc}
\toprule
\multirow{2}{*}{Method}  & \multirow{2}{*}{Model}  & Extra         &  Clean & PGD   & Auto\\ 
       &       & B.P. Steps    &  Acc.     &  100 & Attack  \\ 
\midrule
AT  (\citet{madry2017towards})        & WRN-34-10 & $\checkmark$       & 87.14  & 47.04     & 44.04   \\ 
LAT (\citet{singh2019harnessing}) & WRN-34-10 & $\checkmark$ & 87.80 & 53.04 & 49.12\\
TLA (\citet{mao2019metric}) & WRN-34-10 & $\checkmark$  & 86.21 & 50.03	 & 47.41\\
YOPO (\citet{zhang2019you}) & WRN-34-10 & $\checkmark$  & 87.20 & 47.98	& 44.83\\
Free AT (\citet{shafahi2019adversarial}) & WRN-34-10 & $\checkmark$       & 86.11  & 46.19     & 41.47     \\
TRADES  (\citet{zhang2019theoretically})                      & WRN-34-10 & $\checkmark$       & 84.92  & 56.43     & 53.08   \\ 
MART    (\citet{wang2019improving} )                 & WRN-34-10 & $\checkmark$       & 83.07  &  55.57    & -   \\

AWP    (\citet{wu2020adversarial})               & WRN-34-10 & $\checkmark$       & 85.36  &  59.12    & 56.17   \\
FAT    (\citet{zhang2020attacks})                   & WRN-34-10 & $\checkmark$       & 84.52  &  54.36    & 53.51   \\
Overfitting (\citet{rice2020overfitting}) & WRN-34-20 & $\checkmark$ & 85.34  & 58.00 & 53.42\\
BoT TRADES (\citet{pang2020bag}) & WRN-34-20 & $\checkmark$ & 86.43  & 54.39 & 54.39\\

Adv. Pretraining  (\citet{hendrycks2019pretraining})  & WRN-28-10 & $\checkmark$         & 87.10   &  57.40     & 54.92   \\
\hline
IGAM     (\citet{chan2020thinks} )            & WRN-34-10 & $\checkmark$       & 88.70   & 43.00    & -     \\
RKD     (\citet{goldblum2020adversarially})           & WRN-34-10 & $\times $          & 89.38 & 0.21 & 0  \\
\midrule
Ours (w/o Mixup)             & WRN-34-10 & $\times$           & 91.17$\pm$0.06  & 49.53$\pm$1.44 & 47.38$\pm$1.51      \\
Ours (w/ Mixup)              & WRN-34-10 & $\times$           & 90.76$\pm$0.05  & 56.65$\pm$0.93  & 53.93$\pm$0.77 \\
\bottomrule

\end{tabular}
\end{adjustbox}
\label{tab:rob_transfer}
\end{table}

\textbf{Evaluation of Robustness}: We evaluate MixACM and other methods against standard FGSM and PGD-k as well as the strongest auto-attack by \citet{croce2020reliable}: a parameter-free adversarial attack designed to give a reliable evaluation of robustness. It is an ensemble of four white and black-box attacks. Apart from these, we also report robustness of our robust models on auto pgd-ce~\cite{croce2020reliable}, auto pgd-dlr~\cite{croce2020reliable}, FAB~\cite{croce2020minimally} and query-based black-box attack called Square attack~\cite{andriushchenko2020square}. For PGD-k, we use the standard value of step size ($2/255$), and evaluated the model with an increasing number of iterations ($k$), including an extreme version: PGD-500. 

\textbf{Pre-Trained Robust Models}: For CIFAR experiments, we use WideResNet-28-10 and WideResNet-34-20 \cite{zagoruyko8wide} provided by \citet{gowal2020uncovering}. For ImageNet experiments, we use adversarially trained ResNet-50 provided by \citet{salman2020adversarially}.

\textbf{Training Details}:
For CIFAR experiments, we use WideResNet-34-10 \cite{zagoruyko8wide} as student model following \cite{chan2020thinks}. All student models for our proposed method are trained without adversarial perturbations. We trained them for 200 epochs, using batch size of 128, a learning rate of 0.1, cosine learning rate scheduler \cite{loshchilov10sgdr}, momentum optimizer with weight decay of $0.0005$. For our loss, we use $\alpha_{acm}=5000$. For KD loss, we use temperature value of $\gamma=10$ and $\alpha_{kld}=0.95$ and the value for mixup coefficient is $\alpha_{mixup}=1$ whereas $\lambda \sim Beta(\alpha_{mixup}, \alpha_{mixup})$ following \cite{zhang2017mixup}. ImageNet models are trained for 120 epochs. Additional training details are in the supplementary material.

\subsection{Transferring Robustness without Adversarial Examples}
\label{sec:robustness_transfer}

\textbf{CIFAR-10}: To show the effectiveness of our method for robustness transfer and compare it with state-of-the-art, we follow settings of \cite{chan2020thinks}: WideResNet-34-10 as a student on CIFAR-10. We compare our method with two kinds of defense methods: general variants of adversarial training and three robustness transfer methods. For our method, the average $\pm$ variance of five runs for our method are shown in Table~\ref{tab:rob_transfer}. The robustness of our method is comparable to most state-of-the-arts adversarial training methods while still maintaining significantly higher clean accuracy. 

\textbf{ImageNet}: We also consider high-resolution, large-scale ImageNet dataset~\cite{deng2009imagenet}. This dataset is significantly more challenging for robustness~\cite{xie2019feature, Xie2020intriguing}. The results of our method for ResNet-50 student are shown in Table~\ref{tab:transfer_learning_imagenet}. For our method, we show two results: one with only mixup augmentation and one with mixup and random augmentation \cite{cubuk2020randaugment}. We observed that addition of random augmentation further improves the results of our method. Our method outperforms both Free \cite{shafahi2019adversarial}  and Fast \cite{wong2020fast} AT in accuracy and robustness, significantly. 

\begin{table}[t]
        \centering
       \caption{\textbf{Comparison on ImageNet.} Results of robustness transfer for large-scale ImageNet dataset with ResNet-50 with two different training settings. Our method has significantly better robustness while also achieving comparable or better clean accuracy although it does not require any additional back-propagation steps. }
    \label{tab:transfer_learning_imagenet}
       
        \begin{adjustbox}{width=1\textwidth}
            \begin{tabular}{lccccc|ccccc}
            \toprule
      
              Method&   Train $\epsilon$  & Acc. &  PGD10 & PGD50 & PGD100 &    Train $\epsilon$  & Acc. & PGD10 & PGD50 & PGD100       \\

            \hline
            \multicolumn{6}{c}{\text{Test $\epsilon=2/255$}} &  \multicolumn{5}{|c}{\text{Test $\epsilon=4/255$}}\\
            \hline
            Free-AT~\cite{shafahi2019adversarial}      & 2 & \textbf{64.45}  & 43.52  & 43.39 & 43.40  & 4  & 60.21 &  32.77 & 31.88   & 31.82 \\ 
            Fast-AT~\cite{wong2020fast}               & 2  & 60.90  & -      & 43.46  & -  & 4  & 55.45 & -      & 30.28  & - \\
     
            Ours                                      & 0  & 59.64  & 46.03 & 45.96 & 45.92  & 0  & 59.64 & 32.70  & 31.51  & 31.39  \\
            Ours+RA                                   & 0  & 62.05  & \textbf{48.54} &\textbf{48.45} & \textbf{48.44}  & 0  & \textbf{62.05} & \textbf{34.93}  & \textbf{33.71}  & \textbf{33.63}  \\
            \bottomrule
            \end{tabular}
        \end{adjustbox}
\end{table}

\begin{table}[t]
\centering
\caption{\textbf{Distillation Results.} Performance for robustness distillation on CIFAR-10 and CIFAR-100 datasets. The teacher model is WideResNet-28-10 and student model is WideResNet-16-10. Robustness is reported for diverse set of attacks of smagnitude $\epsilon=8/255$.}
\begin{adjustbox}{width=1\linewidth}
\begin{tabular}{clccccccccc}
\toprule
\multirow{2}{*}{Dataset}  & \multirow{2}{*}{Method} & \multirow{2}{*}{Acc.} & \multirow{2}{*}{FGSM}  & PGD & PGD  & APGD & APGD & \multirow{2}{*}{FAB} & \multirow{2}{*}{Square}   &  Auto \\
 &  &  &  & 20 & 500  & CE & DLR &  &   & Attack \\
\toprule
\multirow{5}{*}{CIFAR-10}
&Natural     & 93.79 & 5.34 & 0 & 0 & 0 & 0 & 0 & 0 & 0 \\
&PGD7-AT     & 83.90 & 53.07 & 47.55 & 47.36 & 47.28 & 44.52 & 44.89 & 53.26 & 44.51\\
&RKD                 & 88.60 & 27.37 & 0.12 & 0 & 0 & 0 & 0 & 7.31 & 0\\
\cmidrule{2-11}
&Ours \small{w/o mixup}& 90.47 & 56.98 & 40.69 & 39.69 & 39.24 & 36.46 & 37.21 & 56.30 & 36.42\\
&Ours                & 89.93 & 59.48 & 48.09 & 47.51 & 47.31 & 43.73 & 44.46 & 60.20 & 43.64 \\
\midrule
\multirow{5}{*}{CIFAR-100}
&Natural                   & 76.91 & 3.17 & 0 & 0 & 0 & 0 & 0 & 0 & 0\\
&PGD7-AT                   & 60.29 & 28.37 & 24.94 & 24.75 & 24.71 & 21.94 & 22.16 & 27.97 & 21.94\\
&RKD                               & 67.19 & 10.27 & 0.23 & 0.16 & 0.04 & 0 & 0 & 3.73 & 0\\
\cmidrule{2-11}
&Ours \small{w/o mixup}                 & 60.81 & 30.13 & 25.38 & 25.12 & 24.93 & 19.51 & 19.83 & 29.74 & 19.50\\
&Ours                              & 59.35 & 31.89 & 28.24 & 28.20 & 27.96 & 22.02 & 22.26 & 31.60 & 22.02\\
\bottomrule
\end{tabular}
\end{adjustbox}
\label{tab:distillation}

\end{table}

\subsection{Robustness Distillation for Model Compression}
\label{sec:robustness_compression}
A benefit of our method is its ability to distill robustness across models. To show this, we conduct two sets of experiments. First, we show robustness distillation from WideResNet-28-10 to WideResNet-16-10 for CIFAR-10 and CIFAR-100 datasets. The results are shown in Table \ref{tab:distillation}. To see the effect of robustness transfer across different models, we also conduct an experiment where we changed the size of the model in terms of depth, the number of channels, or both, e.g., WideResNet-28-10 to WideResNet-16-5. The results are shown in Table \ref{tab:compression}. Our method works well for both larger-to-smaller as well as smaller-to-larger models. However, our method does not perform well when both depth and number of channels are decreased dramatically, e.g., from WideResNet-34-20 to WideResNet-10-1 (student's size is 0.04\%  of teacher's size). To elaborate further, we discuss this in the supplementary material.

\begin{table}
\centering
\caption{\textbf{Model Compression}. Comparison of our method with PGD-7 Adversarial Training for accuracy and robustness on various sizes of models. The robustness is reported for $\ell_\infty$ PGD-20 attack of magnitude $\epsilon=8/255$. Our method works well for robustness transfer from larger to smaller as well as, for smaller to larger models.}
\begin{adjustbox}{width=0.9\linewidth}
\begin{tabular}{@{}lllccccc@{}}
\toprule
\multirow{2}{*}{Distill Type} & \multirow{2}{*}{Teacher}& \multirow{2}{*}{Student} &\multirow{2}{*}{Size Ratio}& \multicolumn{2}{c}{PGD7-AT} & \multicolumn{2}{c}{Ours} \\ 
\cmidrule{5-8}
                 &           &                &       & Acc.  & Rob.  & Acc.  & Rob. \\
\midrule
Depth            & WRN-28-10 & WRN-16-10 &  46.92\%         & 83.90 & 47.55 &  89.93 & 48.09 \\
Channel          & WRN-28-10 & WRN-28-5  & 25.04\%          & 83.90 & 47.74 &  90.26 & 48.75 \\
Channel          & WRN-34-20 & WRN-34-10 & 25.01\%          & 84.30 & 49.72 &  87.26 & 52.83 \\
Depth \& Channel & WRN-28-10 & WRN-16-5  & 11.76\%          & 83.51 & 48.11 &  88.96 & 36.95 \\
Depth \& Channel & WRN-34-20 & WRN-16-10 &  9.227\%         & 83.90 & 47.55 &  86.31 & 47.18 \\
Tiny network     & WRN-28-10 & WRN-10-1  & 0.21\%           & 67.11 & 38.85 &  66.65 & 1.15 \\
Tiny network     & WRN-34-20 & WRN-10-1  & 0.04\%           & 67.11 & 38.85 &  63.19 & 3.75 \\
Same network     & WRN-28-10 & WRN-28-10 & 100\%            & 84.17 & 49.23 &  90.48 & 54.05  \\ 
Same network     & WRN-34-20 & WRN-34-20 & 100\%            & 84.20 & 49.11 &  87.17 & 53.60  \\
\midrule
Up Transfer      & WRN-28-10 & WRN-34-20 & 505.85\%            & 84.20 & 49.11 &  90.67 & 58.36  \\

\bottomrule
\end{tabular}
\end{adjustbox}

\label{tab:compression}
\end{table}

\subsection{Robust Transfer Learning}
\label{sec:robustness_transfer_learning}
\begin{figure}
    \begin{minipage}[h]{0.52\textwidth}
        \centering
        \caption{Robustness Transfer with Smaller Data. Figure shows robustness and accuracy (\%) for the student model trained with different fractions of CIFAR-10 dataset. Our method achieves significantly better clean accuracy and robustness compared with adversarial training. Dotted blue line shows the performance of adversarial training with full data.}
        \includegraphics[width=1\linewidth]{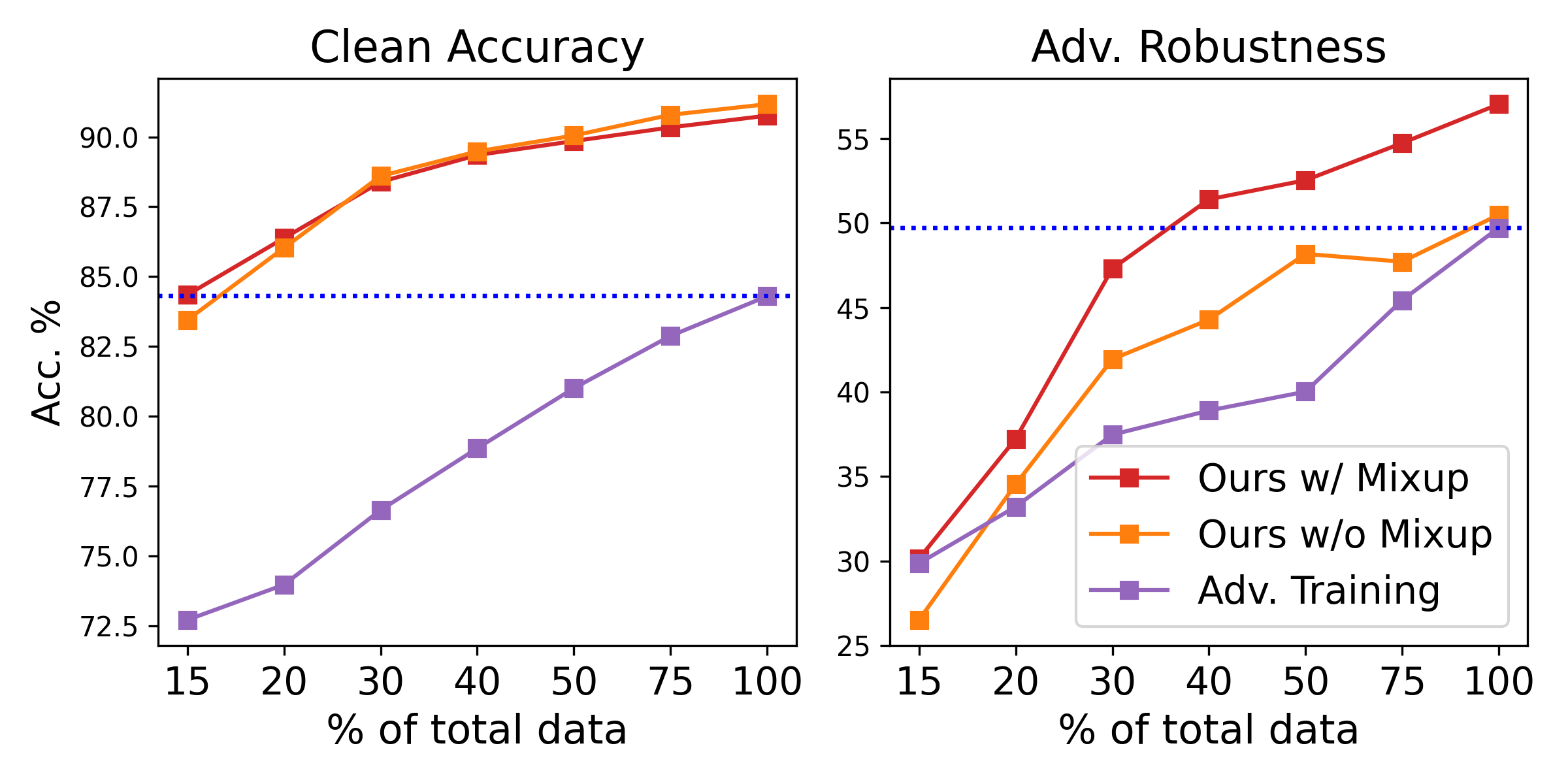}

        \label{fig:tranfer_learning_small_data}
    \end{minipage}
    \quad
    \begin{minipage}[h]{0.47\textwidth}
        \centering
        \captionof{table}{Robustness Transfer from ImageNet to CIFAR-100. Teacher model is adversarially trained on ImageNet and student is trained on CIFAR-100. Our method achieves significantly better clean accuracy and better robustness compared with both adversarial training and IGAM \cite{chan2020thinks}. The robustness is reported according to \cite{chan2020thinks} for a fair comparison. Reported results for our method are mean of five repetitions.}
        \label{tab:transfer_learning}
        \begin{adjustbox}{width=1\textwidth}
        \begin{tabular}{lccccc}
         \toprule
        \multirow{2}{*}{Defense}     &  \multirow{2}{*}{Acc.}    & \multirow{2}{*}{FGSM}     & \multicolumn{3}{c}{PGD-k}\\
        \cmidrule{4-6}
          &  &  & 5  & 10  & 20 \\
        \midrule
        Natural~\cite{chan2020thinks}  & 78.70 & 7.95 & 0.13 & 0.03 & 0  \\
        PGD7-AT~\cite{chan2020thinks} & 60.40 & 29.10 & 29.30 & 24.30 & 23.50 \\
        IGAM~\cite{chan2020thinks}     & 62.39 & 34.31 & 29.59 & 24.05 & 21.74  \\
        \midrule
        Ours    \tiny w/o Mixup       & \textbf{68.91} & 29.09 & 28.65 & 21.70 & 20.27 \\
        Ours                           & 65.69 & 32.09 & \textbf{32.04} & \textbf{25.58} & \textbf{24.14} \\
        \bottomrule
        \end{tabular}
        \end{adjustbox}
    \end{minipage}
\end{figure}

To further show the effectiveness of our method, we conduct two sets of experiments under transfer learning settings: transfer learning from one dataset to another dataset and transfer learning with fewer data samples. For first experiment, we follow \citet{hendrycks2019pretraining}'s setting: our teacher model is a WideResNet-28-10 trained on ImageNet of size 32 $\times$ 32 and student is a WideResNet-34-10. The results for the CIFAR-100 dataset under these settings are shown in Table \ref{tab:transfer_learning}.

For the second part, we show the effectiveness of our method with fewer data points. To this end, we randomly choose a smaller portion of CIFAR-10 and conducted experiments on this smaller portion of data with our method as well with PGD-7 adversarial training. The results are shown in Figure \ref{fig:tranfer_learning_small_data}. Our method outperforms adversarial training significantly for both robustness and clean accuracy. Using just 30\% of the data, our method can match the robustness and clean accuracy of a robust model trained on all the data. \par

\subsection{Ablation Studies}
\textbf{Effect of Individual Components:} To see the effect of individual components of our proposed method, we perform an experiment where we try different settings. The results are shown in Table \ref{tab:ablation}(a). ACM loss alone is sufficient to transfer significant robustness but adding mixup and soft labels improves the transferability. However, the improvement in robustness comes at the cost of clean accuracy. Also, note that only soft labels are not enough to transfer robustness as shown by KD Only column. This is in line with the observations of \citet{goldblum2019adversarially}.

\textbf{Role of Intermediate Features}:
To understand the role of low, mid, and high-level features, we performed experiments on CIFAR-10 by progressively changing blocks used for distillation. For this ablation study, we kept all the standard settings reported in the paper (supplementary material section A.1). Our correspondence of blocks and features is as follows: block 2: low-level features; block 3: mid-level features; block 4: high-level features. Please note that block 1 corresponds to the output of the first layer only. Therefore, we do not call it low-level features.

The results are reported in Table~\ref{tab:ablation} (b). In summary, all level features (low, mid, high level) improve robustness and accuracy. However, mid-level features seem to be more critical for robustness and high-level features for accuracy. For more details, please refer to supplementary material. 

\textbf{Other Ablation Studies}: We have conducted several additional ablation studies and discussed them in the supplementary material (Section A). We recapitulate the purpose of these ablation studies here for reference. First, we conducted a study to understand the effect of direct distillation vs. distillation via MixACM. Second, we performed an ablation study to see the effect of three hyperparameters used by our loss ($\alpha_{acm}, \alpha_{kld}, \gamma$). Third, we performed a study to see how different transforms ($g_c(.)$) effect the results. Fourth, to see the limitation of our method, we performed an experiment with progressively decreasing the size of student. Finally, we also compared activation maps of a normally trained student, teacher and model trained with our method. 

\begin{table*}
\centering
\caption{\textbf{Ablation Studies}: (a) Impact of different components of our proposed distillation method. We use WideResNet-34-10 as student following Table 1 in the main text. (b) Impact of of using different intermediate features on clean accuracy and robustness of student.}
\label{tab:ablation}
\begin{tabular}{cc}
  (a) & (b) \\
    \begin{adjustbox}{width=0.55\linewidth}
    \begin{tabular}[width=1\textwidth]{l|c|c|c|c|c}
    \hline
     & Std.    & w/o & w/o   & Only & Only \\
     & Setting & KD  & Mixup & ACM  & KD \\
         \hline
    Mapping  ($g_c$) & \checkmark & \checkmark & \checkmark & \checkmark &  \\
    Soft Labels                        &\checkmark &   & \checkmark & & \checkmark\\
    Mixup                      & \checkmark & \checkmark &  & &   \\
    ACM Loss                  & \checkmark &\checkmark  & \checkmark & \checkmark &   \\
    \hline
    \hline
    Accuracy              & 90.76 & 92.50 & 91.17 & 92.87 & 89.38 \\
    Robustness            & 56.65 & 52.29 & 49.53 & 48.38 & 0.21\\
    \hline
    \end{tabular}
    \end{adjustbox}

&
    \begin{adjustbox}{width=0.35\linewidth}
    \begin{tabular}{l|cc}
         \hline
         \textbf{Features Used}& \textbf{Acc.} & \textbf{Rob.} \\
      
         \hline
          Low-Level (2)    & 86.36 & 17.24 \\
          Mid-Level (3)    & 88.30  & 39.37 \\
          High-Level (4)   & 91.18 & 33.13 \\
          \hline
          Low+Mid Level (2+3)      & 88.15 & 41.92 \\
          Mid+High Level (3+4)     & 90.69 & 52.79 \\
          
         \hline
         First Layer Only (1)       & 86.00    & 0.31  \\
         No First Layer (2+3+4)    & 90.69 & 56.15 \\
         \hline
         All Features   & 90.76  & 56.65     \\
         \hline
    \end{tabular}
    \end{adjustbox}
    \\

\end{tabular}

\end{table*}

\section{Conclusion}
In this paper, we have investigated robustness transfer from a theoretical and algorithmic perspective. The general principle behind most successful adversarial defense methods is the generation of adversarial examples during training. However, adversarial training is complicated as it requires more data, larger models, higher compute resources and degrades clean accuracy. We investigated whether we can have robustness more efficiently through robustness transfer. To this end, we first presented a theoretical result bounding adversarial loss of student model with the distance between robust teacher and student on mixup examples. We, then, proposed a novel robustness transfer method that is based on channel-wise analysis of robustness. Our method can achieve significant robustness against challenging and reliable auto-attack as well as other standard attacks while outperforming previous work in terms of clean accuracy. The proposed method also works well for the distillation of robustness to a smaller model, and for transfer learning with fewer data points.

\textbf{Acknowledgements}: Authors are thankful to Ferjad Naeem, Dr. Nauman, and anonymous reviewers for their valuable feedback. 

{\small
\bibliographystyle{plainnat}

\bibliography{refs}

\begin{thebibliography}{101}
\providecommand{\natexlab}[1]{#1}
\providecommand{\url}[1]{\texttt{#1}}
\expandafter\ifx\csname urlstyle\endcsname\relax
  \providecommand{\doi}[1]{doi: #1}\else
  \providecommand{\doi}{doi: \begingroup \urlstyle{rm}\Url}\fi

\bibitem[Andriushchenko and Flammarion(2020)]{andriushchenko2020understanding}
Maksym Andriushchenko and Nicolas Flammarion.
\newblock Understanding and improving fast adversarial training.
\newblock \emph{Advances in Neural Information Processing Systems}, 33, 2020.

\bibitem[Andriushchenko et~al.(2020)Andriushchenko, Croce, Flammarion, and
  Hein]{andriushchenko2020square}
Maksym Andriushchenko, Francesco Croce, Nicolas Flammarion, and Matthias Hein.
\newblock Square attack: a query-efficient black-box adversarial attack via
  random search.
\newblock In \emph{European Conference on Computer Vision}, pages 484--501.
  Springer, 2020.

\bibitem[Archambault et~al.(2019)Archambault, Mao, Guo, and
  Zhang]{archambault2019mixup}
Guillaume~P Archambault, Yongyi Mao, Hongyu Guo, and Richong Zhang.
\newblock Mixup as directional adversarial training.
\newblock \emph{arXiv preprint arXiv:1906.06875}, 2019.

\bibitem[Athalye et~al.(2018)Athalye, Carlini, and
  Wagner]{athalye2018obfuscated}
Anish Athalye, Nicholas Carlini, and David Wagner.
\newblock Obfuscated gradients give a false sense of security: Circumventing
  defenses to adversarial examples.
\newblock In \emph{International Conference on Machine Learning}, pages
  274--283. PMLR, 2018.

\bibitem[Awais et~al.(2020)Awais, Shamshad, and Bae]{awais2020towards}
Muhammad Awais, Fahad Shamshad, and Sung-Ho Bae.
\newblock Towards an adversarially robust normalization approach.
\newblock \emph{arXiv preprint arXiv:2006.11007}, 2020.

\bibitem[Awais et~al.(2021)Awais, Zhou, Xu, Hong, Luo, Bae, and
  Li]{awais2021adversarial}
Muhammad Awais, Fengwei Zhou, Hang Xu, Lanqing Hong, Ping Luo, Sung-Ho Bae, and
  Zhenguo Li.
\newblock Adversarial robustness for unsupervised domain adaptation.
\newblock In \emph{Proceedings of the IEEE/CVF International Conference on
  Computer Vision}, pages 8568--8577, 2021.

\bibitem[Bai et~al.(2021)Bai, Zeng, Jiang, Xia, Ma, and Wang]{bai2021improving}
Yang Bai, Yuyuan Zeng, Yong Jiang, Shu-Tao Xia, Xingjun Ma, and Yisen Wang.
\newblock Improving adversarial robustness via channel-wise activation
  suppressing.
\newblock \emph{arXiv preprint arXiv:2103.08307}, 2021.

\bibitem[Bau et~al.(2020)Bau, Zhu, Strobelt, Lapedriza, Zhou, and
  Torralba]{bau2020units}
David Bau, Jun-Yan Zhu, Hendrik Strobelt, Agata Lapedriza, Bolei Zhou, and
  Antonio Torralba.
\newblock Understanding the role of individual units in a deep neural network.
\newblock \emph{Proceedings of the National Academy of Sciences}, 2020.
\newblock ISSN 0027-8424.
\newblock \doi{10.1073/pnas.1907375117}.
\newblock URL \url{https://www.pnas.org/content/early/2020/08/31/1907375117}.

\bibitem[Biggio et~al.(2013)Biggio, Corona, Maiorca, Nelson, {\v{S}}rndi{\'c},
  Laskov, Giacinto, and Roli]{biggio2013evasion}
Battista Biggio, Igino Corona, Davide Maiorca, Blaine Nelson, Nedim
  {\v{S}}rndi{\'c}, Pavel Laskov, Giorgio Giacinto, and Fabio Roli.
\newblock Evasion attacks against machine learning at test time.
\newblock In \emph{Joint European conference on machine learning and knowledge
  discovery in databases}, pages 387--402. Springer, 2013.

\bibitem[Brown et~al.(2017)Brown, Man{\'e}, Roy, Abadi, and
  Gilmer]{brown2017adversarial}
Tom~B Brown, Dandelion Man{\'e}, Aurko Roy, Mart{\'\i}n Abadi, and Justin
  Gilmer.
\newblock Adversarial patch.
\newblock \emph{arXiv preprint arXiv:1712.09665}, 2017.

\bibitem[Buciluǎ et~al.(2006)Buciluǎ, Caruana, and
  Niculescu-Mizil]{bucilu2006model}
Cristian Buciluǎ, Rich Caruana, and Alexandru Niculescu-Mizil.
\newblock Model compression.
\newblock In \emph{Proceedings of the 12th ACM SIGKDD international conference
  on Knowledge discovery and data mining}, pages 535--541, 2006.

\bibitem[Buckman et~al.(2018)Buckman, Roy, Raffel, and
  Goodfellow]{buckman2018thermometer}
Jacob Buckman, Aurko Roy, Colin Raffel, and Ian Goodfellow.
\newblock Thermometer encoding: One hot way to resist adversarial examples.
\newblock In \emph{International Conference on Learning Representations}, 2018.

\bibitem[Bunk et~al.(2021)Bunk, Chattopadhyay, Manjunath, and
  Chandrasekaran]{bunk2021adversarially}
Jason Bunk, Srinjoy Chattopadhyay, BS~Manjunath, and Shivkumar Chandrasekaran.
\newblock Adversarially optimized mixup for robust classification.
\newblock \emph{arXiv preprint arXiv:2103.11589}, 2021.

\bibitem[Carlini and Wagner(2017)]{carlini2017towards}
Nicholas Carlini and David Wagner.
\newblock Towards evaluating the robustness of neural networks.
\newblock In \emph{2017 ieee symposium on security and privacy (sp)}, pages
  39--57. IEEE, 2017.

\bibitem[Carmon et~al.(2019)Carmon, Raghunathan, Schmidt, Liang, and
  Duchi]{carmon2019unlabeled}
Yair Carmon, Aditi Raghunathan, Ludwig Schmidt, Percy Liang, and John~C Duchi.
\newblock Unlabeled data improves adversarial robustness.
\newblock \emph{arXiv preprint arXiv:1905.13736}, 2019.

\bibitem[Chan et~al.(2020)Chan, Tay, and Ong]{chan2020thinks}
Alvin Chan, Yi~Tay, and Yew-Soon Ong.
\newblock What it thinks is important is important: Robustness transfers
  through input gradients.
\newblock In \emph{Proceedings of the IEEE/CVF Conference on Computer Vision
  and Pattern Recognition}, pages 332--341, 2020.

\bibitem[Chen et~al.(2021)Chen, Zhang, Xu, Hu, Niu, Chen, and
  Sugiyama]{chen2021guided}
Chen Chen, Jingfeng Zhang, Xilie Xu, Tianlei Hu, Gang Niu, Gang Chen, and
  Masashi Sugiyama.
\newblock Guided interpolation for adversarial training.
\newblock \emph{arXiv e-prints}, pages arXiv--2102, 2021.

\bibitem[Croce and Hein(2020{\natexlab{a}})]{croce2020minimally}
Francesco Croce and Matthias Hein.
\newblock Minimally distorted adversarial examples with a fast adaptive
  boundary attack.
\newblock In \emph{International Conference on Machine Learning}, pages
  2196--2205. PMLR, 2020{\natexlab{a}}.

\bibitem[Croce and Hein(2020{\natexlab{b}})]{croce2020reliable}
Francesco Croce and Matthias Hein.
\newblock Reliable evaluation of adversarial robustness with an ensemble of
  diverse parameter-free attacks.
\newblock In \emph{ICML}, 2020{\natexlab{b}}.

\bibitem[Cubuk et~al.(2020)Cubuk, Zoph, Shlens, and Le]{cubuk2020randaugment}
Ekin~D Cubuk, Barret Zoph, Jonathon Shlens, and Quoc~V Le.
\newblock Randaugment: Practical automated data augmentation with a reduced
  search space.
\newblock In \emph{Proceedings of the IEEE/CVF Conference on Computer Vision
  and Pattern Recognition Workshops}, pages 702--703, 2020.

\bibitem[Das et~al.(2018)Das, Shanbhogue, Chen, Hohman, Li, Chen, Kounavis, and
  Chau]{das2018compression}
Nilaksh Das, Madhuri Shanbhogue, Shang-Tse Chen, Fred Hohman, Siwei Li,
  Li~Chen, Michael~E Kounavis, and Duen~Horng Chau.
\newblock Compression to the rescue: Defending from adversarial attacks across
  modalities.
\newblock In \emph{ACM SIGKDD Conference on Knowledge Discovery and Data
  Mining}, 2018.

\bibitem[Deng et~al.(2009)Deng, Dong, Socher, Li, Li, and
  Fei-Fei]{deng2009imagenet}
Jia Deng, Wei Dong, Richard Socher, Li-Jia Li, Kai Li, and Li~Fei-Fei.
\newblock Imagenet: A large-scale hierarchical image database.
\newblock In \emph{2009 IEEE conference on computer vision and pattern
  recognition}, pages 248--255. Ieee, 2009.

\bibitem[Dhillon et~al.(2018)Dhillon, Azizzadenesheli, Lipton, Bernstein,
  Kossaifi, Khanna, and Anandkumar]{dhillon2018stochastic}
Guneet~S Dhillon, Kamyar Azizzadenesheli, Zachary~C Lipton, Jeremy~D Bernstein,
  Jean Kossaifi, Aran Khanna, and Animashree Anandkumar.
\newblock Stochastic activation pruning for robust adversarial defense.
\newblock In \emph{International Conference on Learning Representations}, 2018.

\bibitem[Engstrom et~al.(2019)Engstrom, Ilyas, Santurkar, Tsipras, Tran, and
  Madry]{engstrom2019adversarial}
Logan Engstrom, Andrew Ilyas, Shibani Santurkar, Dimitris Tsipras, Brandon
  Tran, and Aleksander Madry.
\newblock Adversarial robustness as a prior for learned representations.
\newblock \emph{arXiv preprint arXiv:1906.00945}, 2019.

\bibitem[Gao and Pavel(2017)]{gao2017properties}
Bolin Gao and Lacra Pavel.
\newblock On the properties of the softmax function with application in game
  theory and reinforcement learning.
\newblock \emph{arXiv preprint arXiv:1704.00805}, 2017.

\bibitem[Goldblum et~al.(2020{\natexlab{a}})Goldblum, Fowl, Feizi, and
  Goldstein]{goldblum2019adversarially}
Micah Goldblum, Liam Fowl, Soheil Feizi, and Tom Goldstein.
\newblock Adversarially robust distillation.
\newblock \emph{Proceedings of the AAAI Conference on Artificial Intelligence},
  2020{\natexlab{a}}.
\newblock \doi{10.1609/aaai.v34i04.5816}.

\bibitem[Goldblum et~al.(2020{\natexlab{b}})Goldblum, Fowl, Feizi, and
  Goldstein]{goldblum2020adversarially}
Micah Goldblum, Liam Fowl, Soheil Feizi, and Tom Goldstein.
\newblock Adversarially robust distillation.
\newblock In \emph{Proceedings of the AAAI Conference on Artificial
  Intelligence}, pages 3996--4003, 2020{\natexlab{b}}.

\bibitem[Goodfellow et~al.(2014)Goodfellow, Shlens, and
  Szegedy]{goodfellow2014explaining}
Ian~J Goodfellow, Jonathon Shlens, and Christian Szegedy.
\newblock Explaining and harnessing adversarial examples.
\newblock \emph{arXiv preprint arXiv:1412.6572}, 2014.

\bibitem[Gowal et~al.(2020)Gowal, Qin, Uesato, Mann, and
  Kohli]{gowal2020uncovering}
Sven Gowal, Chongli Qin, Jonathan Uesato, Timothy Mann, and Pushmeet Kohli.
\newblock Uncovering the limits of adversarial training against norm-bounded
  adversarial examples.
\newblock \emph{arXiv preprint arXiv:2010.03593}, 2020.
\newblock URL \url{https://arxiv.org/pdf/2010.03593}.

\bibitem[Gu and Rigazio(2014)]{gu2014towards}
Shixiang Gu and Luca Rigazio.
\newblock Towards deep neural network architectures robust to adversarial
  examples.
\newblock \emph{arXiv preprint arXiv:1412.5068}, 2014.

\bibitem[Guo et~al.(2017)Guo, Rana, Cisse, and Van
  Der~Maaten]{guo2017countering}
Chuan Guo, Mayank Rana, Moustapha Cisse, and Laurens Van Der~Maaten.
\newblock Countering adversarial images using input transformations.
\newblock \emph{arXiv preprint arXiv:1711.00117}, 2017.

\bibitem[He et~al.(2016)He, Zhang, Ren, and Sun]{he2016deep}
Kaiming He, Xiangyu Zhang, Shaoqing Ren, and Jian Sun.
\newblock Deep residual learning for image recognition.
\newblock In \emph{Proceedings of the IEEE conference on computer vision and
  pattern recognition}, pages 770--778, 2016.

\bibitem[Hendrycks et~al.(2019)Hendrycks, Lee, and
  Mazeika]{hendrycks2019pretraining}
Dan Hendrycks, Kimin Lee, and Mantas Mazeika.
\newblock Using pre-training can improve model robustness and uncertainty.
\newblock \emph{Proceedings of the International Conference on Machine
  Learning}, 2019.

\bibitem[Heo et~al.(2019{\natexlab{a}})Heo, Kim, Yun, Park, Kwak, and
  Choi]{heo2019comprehensive}
Byeongho Heo, Jeesoo Kim, Sangdoo Yun, Hyojin Park, Nojun Kwak, and Jin~Young
  Choi.
\newblock A comprehensive overhaul of feature distillation.
\newblock In \emph{Proceedings of the IEEE/CVF International Conference on
  Computer Vision}, pages 1921--1930, 2019{\natexlab{a}}.

\bibitem[Heo et~al.(2019{\natexlab{b}})Heo, Lee, Yun, and
  Choi]{heo2019knowledge}
Byeongho Heo, Minsik Lee, Sangdoo Yun, and Jin~Young Choi.
\newblock Knowledge transfer via distillation of activation boundaries formed
  by hidden neurons.
\newblock In \emph{Proceedings of the AAAI Conference on Artificial
  Intelligence}, pages 3779--3787, 2019{\natexlab{b}}.

\bibitem[Hinton et~al.(2015)Hinton, Vinyals, and Dean]{hinton2015distilling}
Geoffrey Hinton, Oriol Vinyals, and Jeff Dean.
\newblock Distilling the knowledge in a neural network.
\newblock \emph{arXiv preprint arXiv:1503.02531}, 2015.

\bibitem[Hsu et~al.(2021)Hsu, Ji, Telgarsky, and Wang]{hsu2021generalization}
Daniel Hsu, Ziwei Ji, Matus Telgarsky, and Lan Wang.
\newblock Generalization bounds via distillation.
\newblock \emph{arXiv preprint arXiv:2104.05641}, 2021.

\bibitem[Huang and Wang(2017)]{huang2017like}
Zehao Huang and Naiyan Wang.
\newblock Like what you like: Knowledge distill via neuron selectivity
  transfer.
\newblock \emph{arXiv preprint arXiv:1707.01219}, 2017.

\bibitem[Ilyas et~al.(2019)Ilyas, Santurkar, Engstrom, Tran, and
  Madry]{ilyas2019adversarial}
Andrew Ilyas, Shibani Santurkar, Logan Engstrom, Brandon Tran, and Aleksander
  Madry.
\newblock Adversarial examples are not bugs, they are features.
\newblock \emph{Advances in neural information processing systems}, 32, 2019.

\bibitem[Kim et~al.(2020)Kim, Tack, and Hwang]{kim2020adversarial}
Minseon Kim, Jihoon Tack, and Sung~Ju Hwang.
\newblock Adversarial self-supervised contrastive learning.
\newblock \emph{Advances in Neural Information Processing Systems}, 33, 2020.

\bibitem[Koratana et~al.(2019)Koratana, Kang, Bailis, and
  Zaharia]{koratana2019lit}
Animesh Koratana, Daniel Kang, Peter Bailis, and Matei Zaharia.
\newblock Lit: Learned intermediate representation training for model
  compression.
\newblock In \emph{International Conference on Machine Learning}, pages
  3509--3518. PMLR, 2019.

\bibitem[Kurakin et~al.(2016)Kurakin, Goodfellow, Bengio,
  et~al.]{kurakin2016adversarial}
Alexey Kurakin, Ian Goodfellow, Samy Bengio, et~al.
\newblock Adversarial examples in the physical world, 2016.

\bibitem[Lee et~al.(2020)Lee, Lee, and Yoon]{lee2020adversarial}
Saehyung Lee, Hyungyu Lee, and Sungroh Yoon.
\newblock Adversarial vertex mixup: Toward better adversarially robust
  generalization.
\newblock In \emph{Proceedings of the IEEE/CVF Conference on Computer Vision
  and Pattern Recognition}, pages 272--281, 2020.

\bibitem[Liao et~al.(2018)Liao, Liang, Dong, Pang, Hu, and
  Zhu]{liao2018defense}
Fangzhou Liao, Ming Liang, Yinpeng Dong, Tianyu Pang, Xiaolin Hu, and Jun Zhu.
\newblock Defense against adversarial attacks using high-level representation
  guided denoiser.
\newblock In \emph{Proceedings of the IEEE Conference on Computer Vision and
  Pattern Recognition}, pages 1778--1787, 2018.

\bibitem[Lin et~al.(2020)Lin, Pong~Lau, Levine, Chellappa, and
  Feizi]{lin2020dual}
Wei-An Lin, Chun Pong~Lau, Alexander Levine, Rama Chellappa, and Soheil Feizi.
\newblock Dual manifold adversarial robustness: Defense against lp and non-lp
  adversarial attacks.
\newblock \emph{Advances in Neural Information Processing Systems Foundation
  (NeurIPS)}, 2020.

\bibitem[Liu et~al.(2018)Liu, Liu, Liu, Wang, Jin, and Wen]{liu2018security}
Qi~Liu, Tao Liu, Zihao Liu, Yanzhi Wang, Yier Jin, and Wujie Wen.
\newblock Security analysis and enhancement of model compressed deep learning
  systems under adversarial attacks.
\newblock In \emph{2018 23rd Asia and South Pacific Design Automation
  Conference (ASP-DAC)}, pages 721--726. IEEE, 2018.

\bibitem[Loshchilov and Hutter(2017)]{loshchilov10sgdr}
Ilya Loshchilov and Frank Hutter.
\newblock Sgdr: Stochastic gradient descent with warm restarts.
\newblock \emph{International Conference on Learning Representations},
  10:\penalty0 3, 2017.

\bibitem[Ma et~al.(2021)Ma, Niu, Gu, Wang, Zhao, Bailey, and
  Lu]{ma2021understanding}
Xingjun Ma, Yuhao Niu, Lin Gu, Yisen Wang, Yitian Zhao, James Bailey, and Feng
  Lu.
\newblock Understanding adversarial attacks on deep learning based medical
  image analysis systems.
\newblock \emph{Pattern Recognition}, 110:\penalty0 107332, 2021.

\bibitem[Madaan et~al.(2020)Madaan, Shin, and Hwang]{madaan2019adversarial}
Divyam Madaan, Jinwoo Shin, and Sung~Ju Hwang.
\newblock Adversarial neural pruning with latent vulnerability suppression.
\newblock In \emph{International Conference on Machine Learning}, pages
  6575--6585. PMLR, 2020.

\bibitem[Madry et~al.(2017)Madry, Makelov, Schmidt, Tsipras, and
  Vladu]{madry2017towards}
Aleksander Madry, Aleksandar Makelov, Ludwig Schmidt, Dimitris Tsipras, and
  Adrian Vladu.
\newblock Towards deep learning models resistant to adversarial attacks.
\newblock \emph{arXiv preprint arXiv:1706.06083}, 2017.

\bibitem[Mao et~al.(2019)Mao, Zhong, Yang, Vondrick, and Ray]{mao2019metric}
Chengzhi Mao, Ziyuan Zhong, Junfeng Yang, Carl Vondrick, and Baishakhi Ray.
\newblock Metric learning for adversarial robustness.
\newblock \emph{arXiv preprint arXiv:1909.00900}, 2019.

\bibitem[Mustafa et~al.(2020)Mustafa, Khan, Hayat, Goecke, Shen, and
  Shao]{mustafa2020deeply}
Aamir Mustafa, Salman~H Khan, Munawar Hayat, Roland Goecke, Jianbing Shen, and
  Ling Shao.
\newblock Deeply supervised discriminative learning for adversarial defense.
\newblock \emph{IEEE transactions on pattern analysis and machine
  intelligence}, 2020.

\bibitem[Najafi et~al.(2019)Najafi, Maeda, Koyama, and
  Miyato]{najafi2019robustness}
Amir Najafi, Shin-ichi Maeda, Masanori Koyama, and Takeru Miyato.
\newblock Robustness to adversarial perturbations in learning from incomplete
  data.
\newblock \emph{arXiv preprint arXiv:1905.13021}, 2019.

\bibitem[Nakkiran(2019)]{nakkiran2019adversarial}
Preetum Nakkiran.
\newblock Adversarial robustness may be at odds with simplicity.
\newblock \emph{arXiv preprint arXiv:1901.00532}, 2019.

\bibitem[Pang et~al.(2019)Pang, Xu, and Zhu]{pang2019mixup}
Tianyu Pang, Kun Xu, and Jun Zhu.
\newblock Mixup inference: Better exploiting mixup to defend adversarial
  attacks.
\newblock \emph{arXiv preprint arXiv:1909.11515}, 2019.

\bibitem[Pang et~al.(2020{\natexlab{a}})Pang, Yang, Dong, Su, and
  Zhu]{pang2020bag}
Tianyu Pang, Xiao Yang, Yinpeng Dong, Hang Su, and Jun Zhu.
\newblock Bag of tricks for adversarial training.
\newblock \emph{arXiv preprint arXiv:2010.00467}, 2020{\natexlab{a}}.

\bibitem[Pang et~al.(2020{\natexlab{b}})Pang, Yang, Dong, Xu, Zhu, and
  Su]{pang2020boosting}
Tianyu Pang, Xiao Yang, Yinpeng Dong, Kun Xu, Jun Zhu, and Hang Su.
\newblock Boosting adversarial training with hypersphere embedding.
\newblock \emph{arXiv preprint arXiv:2002.08619}, 2020{\natexlab{b}}.

\bibitem[Papernot et~al.(2016)Papernot, McDaniel, Wu, Jha, and
  Swami]{papernot2016distillation}
Nicolas Papernot, Patrick McDaniel, Xi~Wu, Somesh Jha, and Ananthram Swami.
\newblock Distillation as a defense to adversarial perturbations against deep
  neural networks.
\newblock In \emph{2016 IEEE symposium on security and privacy (SP)}, pages
  582--597. IEEE, 2016.

\bibitem[Papernot et~al.(2017)Papernot, McDaniel, Goodfellow, Jha, Celik, and
  Swami]{papernot2017practical}
Nicolas Papernot, Patrick McDaniel, Ian Goodfellow, Somesh Jha, Z~Berkay Celik,
  and Ananthram Swami.
\newblock Practical black-box attacks against machine learning.
\newblock In \emph{Proceedings of the 2017 ACM on Asia conference on computer
  and communications security}, pages 506--519, 2017.

\bibitem[Rakin et~al.(2018)Rakin, Yi, Gong, and Fan]{rakin2018defend}
Adnan~Siraj Rakin, Jinfeng Yi, Boqing Gong, and Deliang Fan.
\newblock Defend deep neural networks against adversarial examples via fixed
  and dynamic quantized activation functions.
\newblock \emph{arXiv preprint arXiv:1807.06714}, 2018.

\bibitem[Rice et~al.(2020)Rice, Wong, and Kolter]{rice2020overfitting}
Leslie Rice, Eric Wong, and Zico Kolter.
\newblock Overfitting in adversarially robust deep learning.
\newblock In \emph{International Conference on Machine Learning}, pages
  8093--8104. PMLR, 2020.

\bibitem[Romero et~al.(2014)Romero, Ballas, Kahou, Chassang, Gatta, and
  Bengio]{romero2014fitnets}
Adriana Romero, Nicolas Ballas, Samira~Ebrahimi Kahou, Antoine Chassang, Carlo
  Gatta, and Yoshua Bengio.
\newblock Fitnets: Hints for thin deep nets.
\newblock \emph{arXiv preprint arXiv:1412.6550}, 2014.

\bibitem[Ross and Doshi-Velez(2018)]{ross2018improving}
Andrew Ross and Finale Doshi-Velez.
\newblock Improving the adversarial robustness and interpretability of deep
  neural networks by regularizing their input gradients.
\newblock In \emph{Proceedings of the AAAI Conference on Artificial
  Intelligence}, 2018.

\bibitem[Salman et~al.(2020)Salman, Ilyas, Engstrom, Kapoor, and
  Madry]{salman2020adversarially}
Hadi Salman, Andrew Ilyas, Logan Engstrom, Ashish Kapoor, and Aleksander Madry.
\newblock Do adversarially robust imagenet models transfer better?
\newblock \emph{arXiv preprint arXiv:2007.08489}, 2020.

\bibitem[Schmidt et~al.(2018)Schmidt, Santurkar, Tsipras, Talwar, and
  M{\k{a}}dry]{schmidt2018adversarially}
Ludwig Schmidt, Shibani Santurkar, Dimitris Tsipras, Kunal Talwar, and
  Aleksander M{\k{a}}dry.
\newblock Adversarially robust generalization requires more data.
\newblock \emph{arXiv preprint arXiv:1804.11285}, 2018.

\bibitem[Sehwag et~al.(2020)Sehwag, Wang, Mittal, and Jana]{sehwag2020hydra}
Vikash Sehwag, Shiqi Wang, Prateek Mittal, and Suman Jana.
\newblock Hydra: Pruning adversarially robust neural networks.
\newblock \emph{Advances in Neural Information Processing Systems (NeurIPS)},
  7, 2020.

\bibitem[Shafahi et~al.(2019{\natexlab{a}})Shafahi, Najibi, Ghiasi, Xu,
  Dickerson, Studer, Davis, Taylor, and Goldstein]{shafahi2019adversarial}
Ali Shafahi, Mahyar Najibi, Amin Ghiasi, Zheng Xu, John Dickerson, Christoph
  Studer, Larry~S Davis, Gavin Taylor, and Tom Goldstein.
\newblock Adversarial training for free!
\newblock \emph{arXiv preprint arXiv:1904.12843}, 2019{\natexlab{a}}.

\bibitem[Shafahi et~al.(2019{\natexlab{b}})Shafahi, Saadatpanah, Zhu, Ghiasi,
  Studer, Jacobs, and Goldstein]{shafahi2019adversarially}
Ali Shafahi, Parsa Saadatpanah, Chen Zhu, Amin Ghiasi, Christoph Studer, David
  Jacobs, and Tom Goldstein.
\newblock Adversarially robust transfer learning.
\newblock In \emph{International Conference on Learning Representations},
  2019{\natexlab{b}}.

\bibitem[Shen et~al.(2019)Shen, He, and Xue]{shen2019meal}
Zhiqiang Shen, Zhankui He, and Xiangyang Xue.
\newblock Meal: Multi-model ensemble via adversarial learning.
\newblock In \emph{Proceedings of the AAAI Conference on Artificial
  Intelligence}, volume~33, pages 4886--4893, 2019.

\bibitem[Singh et~al.(2019)Singh, Sinha, Kumari, Machiraju, Krishnamurthy, and
  Balasubramanian]{singh2019harnessing}
Mayank Singh, Abhishek Sinha, Nupur Kumari, Harshitha Machiraju, Balaji
  Krishnamurthy, and Vineeth~N Balasubramanian.
\newblock Harnessing the vulnerability of latent layers in adversarially
  trained models.
\newblock \emph{arXiv preprint arXiv:1905.05186}, 2019.

\bibitem[Srinivas and Fleuret(2018)]{srinivas2018knowledge}
Suraj Srinivas and Fran{\c{c}}ois Fleuret.
\newblock Knowledge transfer with jacobian matching.
\newblock In \emph{International Conference on Machine Learning}, pages
  4723--4731. PMLR, 2018.

\bibitem[Szegedy et~al.(2013)Szegedy, Zaremba, Sutskever, Bruna, Erhan,
  Goodfellow, and Fergus]{szegedy2013intriguing}
Christian Szegedy, Wojciech Zaremba, Ilya Sutskever, Joan Bruna, Dumitru Erhan,
  Ian Goodfellow, and Rob Fergus.
\newblock Intriguing properties of neural networks.
\newblock \emph{arXiv preprint arXiv:1312.6199}, 2013.

\bibitem[Tian et~al.(2019)Tian, Krishnan, and Isola]{tian2019contrastive}
Yonglong Tian, Dilip Krishnan, and Phillip Isola.
\newblock Contrastive representation distillation.
\newblock In \emph{International Conference on Learning Representations}, 2019.

\bibitem[Tram{\`e}r et~al.(2017)Tram{\`e}r, Kurakin, Papernot, Goodfellow,
  Boneh, and McDaniel]{tramer2017ensemble}
Florian Tram{\`e}r, Alexey Kurakin, Nicolas Papernot, Ian Goodfellow, Dan
  Boneh, and Patrick McDaniel.
\newblock Ensemble adversarial training: Attacks and defenses.
\newblock \emph{arXiv preprint arXiv:1705.07204}, 2017.

\bibitem[Tsipras et~al.(2018)Tsipras, Santurkar, Engstrom, Turner, and
  Madry]{tsipras2018robustness}
Dimitris Tsipras, Shibani Santurkar, Logan Engstrom, Alexander Turner, and
  Aleksander Madry.
\newblock Robustness may be at odds with accuracy.
\newblock \emph{arXiv preprint arXiv:1805.12152}, 2018.

\bibitem[Tung and Mori(2019)]{tung2019similarity}
Frederick Tung and Greg Mori.
\newblock Similarity-preserving knowledge distillation.
\newblock In \emph{Proceedings of the IEEE International Conference on Computer
  Vision}, pages 1365--1374, 2019.

\bibitem[Uesato et~al.(2019)Uesato, Alayrac, Huang, Stanforth, Fawzi, and
  Kohli]{uesato2019labels}
Jonathan Uesato, Jean-Baptiste Alayrac, Po-Sen Huang, Robert Stanforth,
  Alhussein Fawzi, and Pushmeet Kohli.
\newblock Are labels required for improving adversarial robustness?
\newblock \emph{arXiv preprint arXiv:1905.13725}, 2019.

\bibitem[Wang et~al.(2018)Wang, Lin, Zhu, Yin, Bertozzi, and
  Osher]{wang2018adversarial}
Bao Wang, Alex~T Lin, Wei Zhu, Penghang Yin, Andrea~L Bertozzi, and Stanley~J
  Osher.
\newblock Adversarial defense via data dependent activation function and total
  variation minimization.
\newblock \emph{arXiv preprint arXiv:1809.08516}, 2018.

\bibitem[Wang et~al.(2019)Wang, Zou, Yi, Bailey, Ma, and Gu]{wang2019improving}
Yisen Wang, Difan Zou, Jinfeng Yi, James Bailey, Xingjun Ma, and Quanquan Gu.
\newblock Improving adversarial robustness requires revisiting misclassified
  examples.
\newblock In \emph{International Conference on Learning Representations}, 2019.

\bibitem[Wong et~al.(2020)Wong, Rice, and Kolter]{wong2020fast}
Eric Wong, Leslie Rice, and J~Zico Kolter.
\newblock Fast is better than free: Revisiting adversarial training.
\newblock \emph{arXiv preprint arXiv:2001.03994}, 2020.

\bibitem[Wu et~al.(2020{\natexlab{a}})Wu, Chen, Cai, He, and Gu]{wu2020does}
Boxi Wu, Jinghui Chen, Deng Cai, Xiaofei He, and Quanquan Gu.
\newblock Does network width really help adversarial robustness?
\newblock \emph{arXiv preprint arXiv:2010.01279}, 2020{\natexlab{a}}.

\bibitem[Wu et~al.(2020{\natexlab{b}})Wu, Xia, and Wang]{wu2020adversarial}
Dongxian Wu, Shu-Tao Xia, and Yisen Wang.
\newblock Adversarial weight perturbation helps robust generalization.
\newblock \emph{Advances in Neural Information Processing Systems}, 33,
  2020{\natexlab{b}}.

\bibitem[Xiao et~al.(2019)Xiao, Zhong, and Zheng]{xiao2019resisting}
Chang Xiao, Peilin Zhong, and Changxi Zheng.
\newblock Resisting adversarial attacks by k-winners-take-all.
\newblock \emph{arXiv preprint arXiv:1905.10510}, 1\penalty0 (2), 2019.

\bibitem[Xie and Yuille(2020)]{Xie2020intriguing}
Cihang Xie and Alan Yuille.
\newblock Intriguing properties of adversarial training at scale.
\newblock In \emph{ICLR}, 2020.

\bibitem[Xie et~al.(2019)Xie, Wu, Maaten, Yuille, and He]{xie2019feature}
Cihang Xie, Yuxin Wu, Laurens van~der Maaten, Alan~L Yuille, and Kaiming He.
\newblock Feature denoising for improving adversarial robustness.
\newblock In \emph{Proceedings of the IEEE/CVF Conference on Computer Vision
  and Pattern Recognition}, pages 501--509, 2019.

\bibitem[Xie et~al.(2020)Xie, Tan, Gong, Yuille, and Le]{xie2020smooth}
Cihang Xie, Mingxing Tan, Boqing Gong, Alan Yuille, and Quoc~V Le.
\newblock Smooth adversarial training.
\newblock \emph{arXiv preprint arXiv:2006.14536}, 2020.

\bibitem[Xu et~al.(2019)Xu, Liu, Zhang, Sun, Zhao, Fan, Gan, and
  Lin]{xu2019interpreting}
Kaidi Xu, Sijia Liu, Gaoyuan Zhang, Mengshu Sun, Pu~Zhao, Quanfu Fan, Chuang
  Gan, and Xue Lin.
\newblock Interpreting adversarial examples by activation promotion and
  suppression.
\newblock \emph{arXiv preprint arXiv:1904.02057}, 2019.

\bibitem[Ye et~al.(2019)Ye, Xu, Liu, Cheng, Lambrechts, Zhang, Zhou, Ma, Wang,
  and Lin]{ye2019adversarial}
Shaokai Ye, Kaidi Xu, Sijia Liu, Hao Cheng, Jan-Henrik Lambrechts, Huan Zhang,
  Aojun Zhou, Kaisheng Ma, Yanzhi Wang, and Xue Lin.
\newblock Adversarial robustness vs. model compression, or both?
\newblock In \emph{Proceedings of the IEEE/CVF International Conference on
  Computer Vision}, pages 111--120, 2019.

\bibitem[Yim et~al.(2017)Yim, Joo, Bae, and Kim]{yim2017gift}
Junho Yim, Donggyu Joo, Jihoon Bae, and Junmo Kim.
\newblock A gift from knowledge distillation: Fast optimization, network
  minimization and transfer learning.
\newblock In \emph{Proceedings of the IEEE Conference on Computer Vision and
  Pattern Recognition}, pages 4133--4141, 2017.

\bibitem[Zagoruyko and Komodakis(2016{\natexlab{a}})]{zagoruyko2016paying}
Sergey Zagoruyko and Nikos Komodakis.
\newblock Paying more attention to attention: Improving the performance of
  convolutional neural networks via attention transfer.
\newblock \emph{arXiv preprint arXiv:1612.03928}, 2016{\natexlab{a}}.

\bibitem[Zagoruyko and Komodakis(2016{\natexlab{b}})]{zagoruyko8wide}
Sergey Zagoruyko and Nikos Komodakis.
\newblock Wide residual networks.
\newblock \emph{NIN}, 8:\penalty0 35--67, 2016{\natexlab{b}}.

\bibitem[Zhai et~al.(2019)Zhai, Cai, He, Dan, He, Hopcroft, and
  Wang]{zhai2019adversarially}
Runtian Zhai, Tianle Cai, Di~He, Chen Dan, Kun He, John Hopcroft, and Liwei
  Wang.
\newblock Adversarially robust generalization just requires more unlabeled
  data.
\newblock \emph{arXiv preprint arXiv:1906.00555}, 2019.

\bibitem[Zhang et~al.(2016)Zhang, Bengio, Hardt, Recht, and
  Vinyals]{zhang2016understanding}
Chiyuan Zhang, Samy Bengio, Moritz Hardt, Benjamin Recht, and Oriol Vinyals.
\newblock Understanding deep learning requires rethinking generalization.
\newblock \emph{arXiv preprint arXiv:1611.03530}, 2016.

\bibitem[Zhang et~al.(2019{\natexlab{a}})Zhang, Zhang, Lu, Zhu, and
  Dong]{zhang2019you}
Dinghuai Zhang, Tianyuan Zhang, Yiping Lu, Zhanxing Zhu, and Bin Dong.
\newblock You only propagate once: Accelerating adversarial training via
  maximal principle.
\newblock \emph{arXiv preprint arXiv:1905.00877}, 2019{\natexlab{a}}.

\bibitem[Zhang et~al.(2019{\natexlab{b}})Zhang, Yu, Jiao, Xing, El~Ghaoui, and
  Jordan]{zhang2019theoretically}
Hongyang Zhang, Yaodong Yu, Jiantao Jiao, Eric Xing, Laurent El~Ghaoui, and
  Michael Jordan.
\newblock Theoretically principled trade-off between robustness and accuracy.
\newblock In \emph{International Conference on Machine Learning}, pages
  7472--7482. PMLR, 2019{\natexlab{b}}.

\bibitem[Zhang et~al.(2017)Zhang, Cisse, Dauphin, and
  Lopez-Paz]{zhang2017mixup}
Hongyi Zhang, Moustapha Cisse, Yann~N Dauphin, and David Lopez-Paz.
\newblock mixup: Beyond empirical risk minimization.
\newblock \emph{arXiv preprint arXiv:1710.09412}, 2017.

\bibitem[Zhang et~al.(2020{\natexlab{a}})Zhang, Xu, Han, Niu, Cui, Sugiyama,
  and Kankanhalli]{zhang2020attacks}
Jingfeng Zhang, Xilie Xu, Bo~Han, Gang Niu, Lizhen Cui, Masashi Sugiyama, and
  Mohan Kankanhalli.
\newblock Attacks which do not kill training make adversarial learning
  stronger.
\newblock In \emph{International Conference on Machine Learning}, pages
  11278--11287. PMLR, 2020{\natexlab{a}}.

\bibitem[Zhang et~al.(2020{\natexlab{b}})Zhang, Zhu, Niu, Han, Sugiyama, and
  Kankanhalli]{zhang2020geometry}
Jingfeng Zhang, Jianing Zhu, Gang Niu, Bo~Han, Masashi Sugiyama, and Mohan
  Kankanhalli.
\newblock Geometry-aware instance-reweighted adversarial training.
\newblock \emph{arXiv preprint arXiv:2010.01736}, 2020{\natexlab{b}}.

\bibitem[Zhang et~al.(2020{\natexlab{c}})Zhang, Deng, Kawaguchi, Ghorbani, and
  Zou]{zhang2020does}
Linjun Zhang, Zhun Deng, Kenji Kawaguchi, Amirata Ghorbani, and James Zou.
\newblock How does mixup help with robustness and generalization?
\newblock \emph{arXiv preprint arXiv:2010.04819}, 2020{\natexlab{c}}.

\bibitem[Zhang and Zhu(2019)]{zhang2019interpreting}
Tianyuan Zhang and Zhanxing Zhu.
\newblock Interpreting adversarially trained convolutional neural networks.
\newblock In \emph{International Conference on Machine Learning}, pages
  7502--7511. PMLR, 2019.

\bibitem[Zhu et~al.(2021)Zhu, Ma, Sun, Chen, Jiang, Chen, and
  Li]{zhu2021towards}
Yao Zhu, Jiacheng Ma, Jiacheng Sun, Zewei Chen, Rongxin Jiang, Yaowu Chen, and
  Zhenguo Li.
\newblock Towards understanding the generative capability of adversarially
  robust classifiers.
\newblock In \emph{Proceedings of the IEEE/CVF International Conference on
  Computer Vision}, pages 7728--7737, 2021.

\end{thebibliography}
}

\newpage
\appendix

\part*{Supplementary Material}


\section{Experimental Results}

\subsection{Additional Details of Experimental Setup}

\textbf{Baselines:} For fair comparisons with other methods, we either use the best results reported in the paper or retrained models with optimal hyper-parameters described by the papers. Details of comparison for all the experimental results in the main text are as follows. 
For Table 1, we take clean accuracy and auto-attack results from \cite{croce2020minimally} and PGD-100 results are the best PGD attack reported results (with the same or similar setting as ours) taken from the respective papers. 
For Table 2, we take \citet{shafahi2019adversarial, wong2020fast}'s reported results and evaluated our model with the same settings of PGD attack. 
For Table 3, we train the same models with PGD7-AT \cite{madry2017towards}, RKD \cite{goldblum2019adversarially} and our method. For PGD7-AT and RKD, we use optimal hyper-parameters reported in the papers.  
For Table 4, we train the same models with PGD7-AT \cite{madry2017towards}, and our method and evaluated all models with the same settings of PGD attack. The size ratio is computed based on the number of trainable parameters of the student to the trainable parameters of teacher.  
The purpose of Table 5 is a comparison of our method with IGAM \cite{chan2020thinks} under transfer learning settings. We used PGD7-AT and normal results reported by them. Unlike them, however, results of our method are mean of five repetitions.  
For Figure 3, we train the same models with the same proportion of data with PGD7-AT and our method.

\textbf{Adversarial Training:} We use standard PGD-7 Adversarial Training with the step size of $2/255$ and $\epsilon=8/255$.

\textbf{Teacher Models:} We use four teachers in the paper. For most of our CIFAR experiments, we use a WideResNet-28-10 trained by \citet{gowal2020uncovering}. For some experiments, we also use WideResNet-34-20 trained on CIFAR-10 by \citet{gowal2020uncovering} and WideResNet-28-10 trained on tiny ImageNet trained by \cite{hendrycks2019pretraining}. For ImageNet experiments, we use ResNet-50 provided by \cite{salman2020adversarially}.

\textbf{Student Models:} Student models share all the architectural design of respective teachers e.g. if teacher model by \citet{gowal2020uncovering} uses Swish activation function, student models also uses Swish activation. For main CIFAR-10 experiments, we use WideResNet-34-10 following many relevant works like \citet{zhang2019theoretically, chan2020thinks, pang2020bag}, etc. We also use students with different widths (20, 10, 5, 1) and depths (34, 28, 22, 16, 10). For ImageNet, we use ResNet50 following \cite{shafahi2019adversarial, wong2020fast}.

\textbf{Optimizer Setting:} For CIFAR experiments, we use SGD with Nesterov momentum 0.9, initial learning rate of 0.1, cosine annealing learning rate decay without restarts, weight decay of $5\times 10^{-4}$ and a batch size of 128. For ImageNet, the model is trained for 120 epochs by SGD with a momentum 0.9, weight decay of $5\times 10^{-5}$, batch size of 2048, initial learning rate of 0.8, and cosine learning rate decay. We also use a gradual warmup strategy that increases the learning rate from 0.16 to 0.8 linearly in the first 5 epochs.

\textbf{Augmentation:} All the experiments of our method use Mixup augmentation with coefficient 1 for CIFAR and 0.2 for ImageNet; unless mentioned otherwise. For CIFAR, we also use standard augmentation: randomly cropping a part of $32 \times 32$ from the padded image followed by a random horizontal flip provided by PyTorch. For ImageNet, we do two experiments: one with only mixup and one with mixup and random augmentation.

\textbf{Hyper-parameter Selection:} Our method adds a new hyper-parameter $\alpha_{acm}$. Two other hyper-parameters of our loss are $\alpha_{kld}$ and temperature $\gamma$. The selection process for them is detailed in Section \ref{sec:ablation_studies}. For $\alpha_{acm}$, we use ablation study  to find optimal range of $\alpha_{acm}$ and all of the other experiments are done with $\alpha_{acm} \in \{2000, 5000\}$ and best results are reported. 

\textbf{Compute Infrastructure:} We train CIFAR models on one NVIDIA Tesla V100. For ImageNet, we train models in a distributed fashion using 32 GPUs in the cloud.

\newpage
\subsection{Activated Channel Maps}
\begin{wrapfigure}{r}{0.25\textwidth}
    \includegraphics[width=0.25\textwidth]{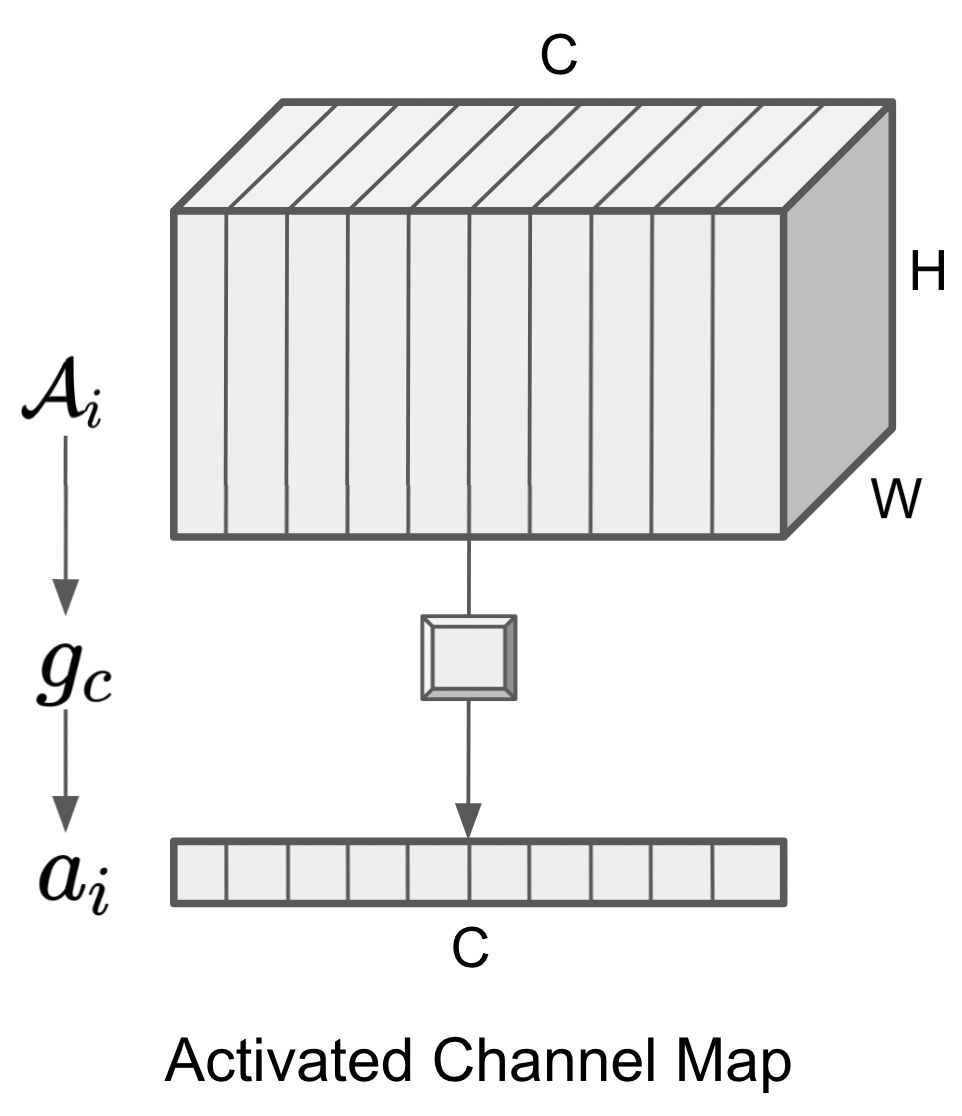}
    \caption{Mapping function $g_c$ is applied channel-wise to get the Activated Channel Maps (ACM).}
    \label{fig:phi_c_supp}
\end{wrapfigure} 
Our method transfers robustness by matching Activated Channel Maps (ACMs) of robust teacher and student on natural examples. To get Activated Channel Maps, we first get an output of a model at a specific layer called activations $\mathcal{A}$. These  activations are then passed through a mapping function to get activation maps: $a_i = g_c(\mathcal{A}_i)$ and normalized with the magnitude. This process is illustrated in Figure \ref{fig:phi_c_supp}. The size of activated channel map is equal to number of channels in a layer e.g. if activation has a size of $\mathcal{A}_i \in \mathbb{R}^{C \times H \times W}$ then activated channel maps shape will be $a_i \in \mathbb{R}^{C \times 1 \times 1}$.

\begin{figure}[t]
    \centering
    \includegraphics[width=1\textwidth]{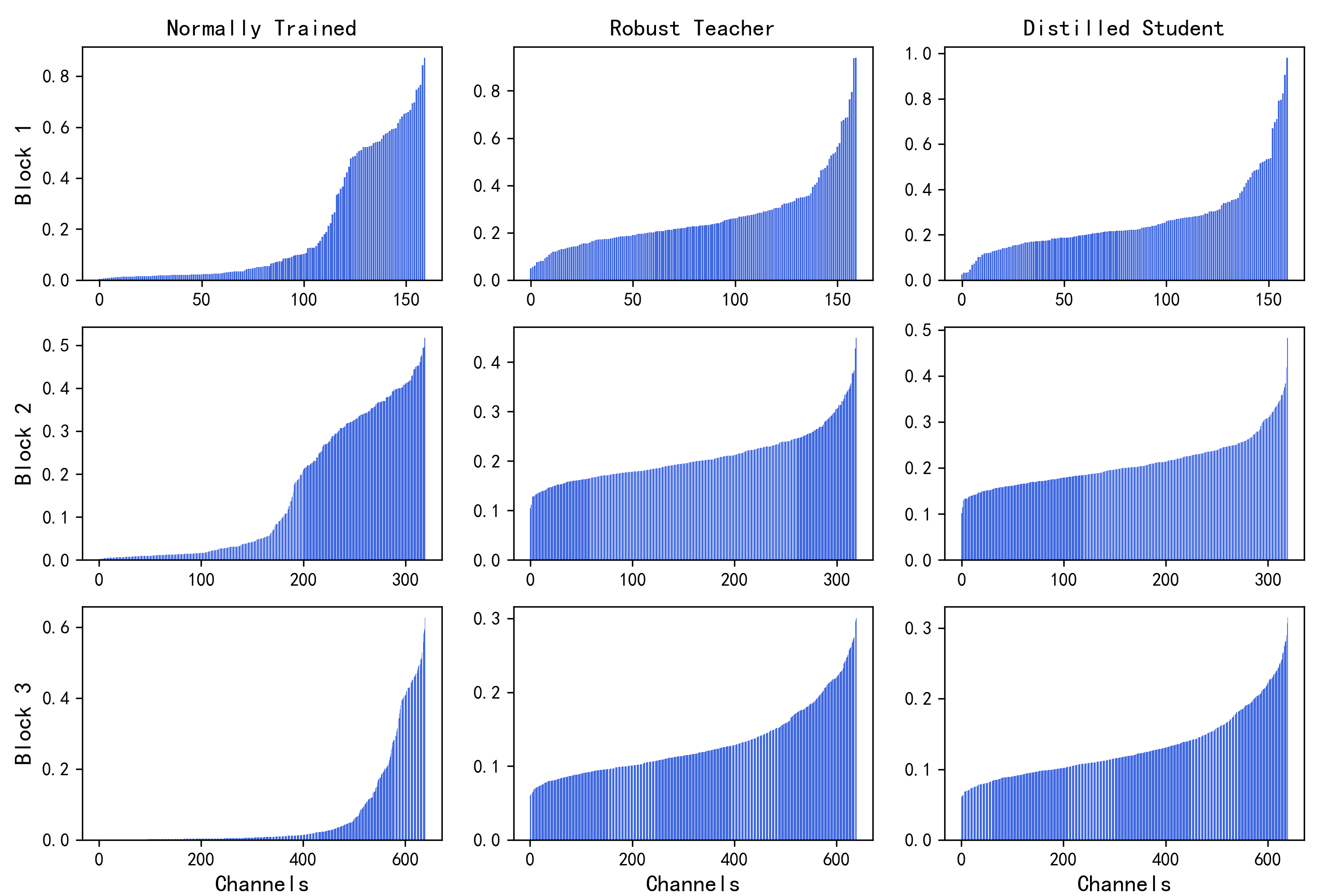}
    \caption{\textbf{Comparison of Activated Channel Maps.} Comparison of Activated Channel Maps of a normally trained model, robust teacher and student trained with the proposed method. The $x$-axis of figure represents channel number of a specific layer and $y$-axis represents ACM value. Our method makes ACM of student similar to the robust teacher.}
    \label{fig:activated_channel_maps}
\end{figure}

We show Activated Channel Maps of three different blocks of a normally trained WideResNet-16-10, a robust teacher WideResNet-28-10 and a student WideResNet-16-10 trained with our method in Figure \ref{fig:activated_channel_maps}. The maps are generated on 500 natural examples of CIFAR-10 test set. The Figure shows an average of maps produced on all the examples and sorted in ascending order. The $x$-axis in the Figure represents channels and the $y$-axis represents the Activated Channel Map values. The distribution of Activated Channel Maps of our method is spectacularly similar to the robust teacher.

\subsection{Limitation of Proposed Method for Distillation}
\begin{figure}
\centering
\begin{subfigure}[b]{1\textwidth}
    \centering
    \includegraphics[width=1\textwidth]{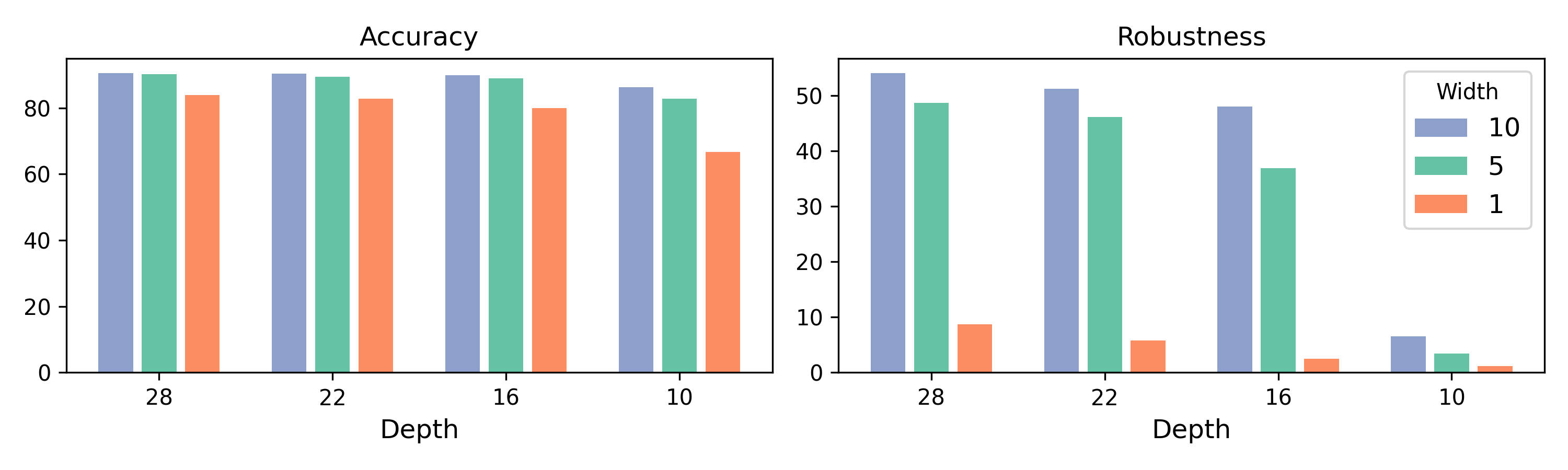}
    \caption{The figure shows effect of decreasing number of layers and number of channels on accuracy and robustness for our method. The teacher model is a robust WideResNet-28-10 and student models are WideResNet-\textit{Depth}-\textit{Width}.}
    \label{fig:ablation_study_compression1}
\end{subfigure}

\hfill
\begin{subfigure}[b]{1\textwidth}
    \centering
    \includegraphics[width=1\textwidth]{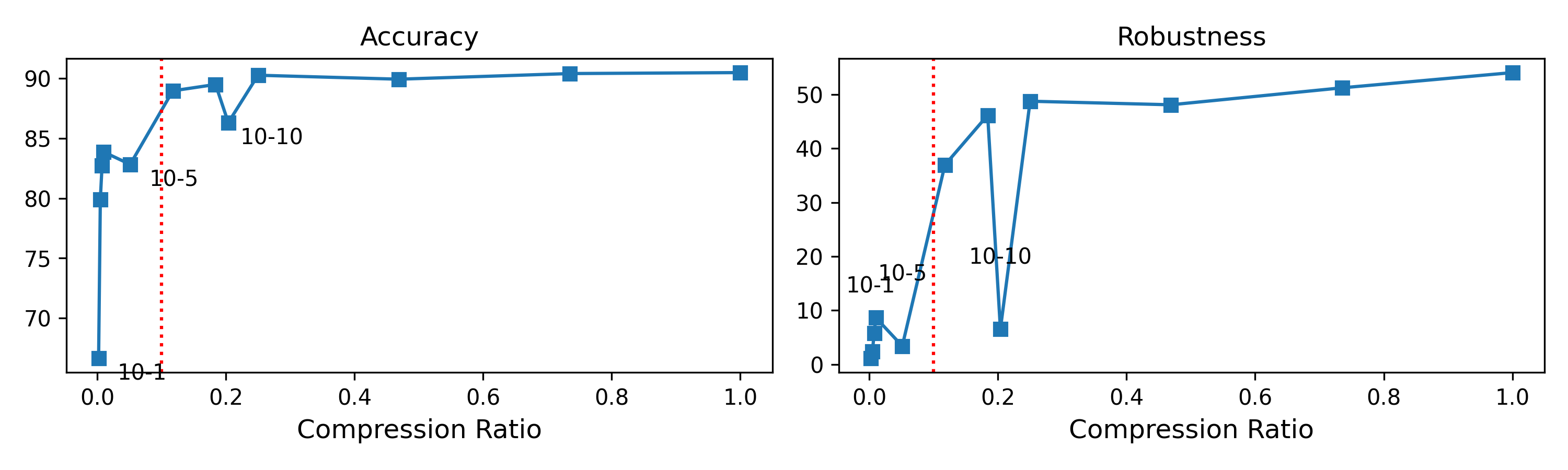}
    \caption{The figure shows compression ration vs accuracy and robustness for our method. The dotted red line represents compression ratio of 0.1. Our method works in transferring robustness beyond this point except for dips caused by WideResNet of depth 10.}
    \label{fig:ablation_study_compression2}
\end{subfigure}
    \caption{\textbf{Limits of Distillation.} Plots show limits of our method in distilling a large teacher model's robustness to a small student model without adversarial training.}
    \label{fig:distillation_limit}

\end{figure}

We show results for distillation with different settings in the paper. To see the limits of our distillation method, we perform an experiment where we progressively reduce the number of channels and layers of the student. The results are shown in Figure \ref{fig:distillation_limit}. For individual values and comparison with PGD7-AT, please see Table \ref{tab:distillation_supp}. 

We observe that our method works well for a compression ratio of $> 0.1$ (student has 0.1$\times$ teacher's trainable parameters). We also observe that robustness transfer deteriorates significantly if depth or width is reduced significantly (e.g., depth reduced from 28 to 10; width reduced from from 10 to 1). The degradation based on width can be attributed to the transformation function and we expect that other functions may work better. The depth degradation may be due to the representational capability of the student model. We leave further investigation of this for future work.

\subsection{Ablation Studies}
\label{sec:ablation_studies}

\subsubsection{Effect of Individual Components} 

To see the effect of individual components of our proposed method, we perform an experiment where we switch different components. The results are shown in Table~\ref{tab:abalation_supp}. 

In summary, ACM loss alone can transfer significant robustness from the teacher; the addition of soft labels and mixup further improves this transfer. Specifically, robustness with only ACM loss is 48.38\%, the addition of soft-labels improves it to 49.53\%, the addition of mixup improves it to 52.29\%, and the addition of both of these components make final robustness to 56.65\%. Also, note that only soft labels are not enough to transfer robustness in this case, as shown by KD Only column. This is in line with the observations of~\citet{goldblum2019adversarially}.

\begin{table}
\centering
\caption{\textbf{Impact of Individual Components.} Impact of different components of our proposed distillation method. We use WideResNet-34-10 as student following Table 1 in the main text. }
\label{tab:abalation_supp}
\begin{tabular}[width=1\textwidth]{l|c|c|c|c|c}
\hline
 & Std. Setting    & w/o KD & w/o Mixup  & Only ACM & Only KD \\
\hline

Mapping Function ($g_c$) & \checkmark & \checkmark & \checkmark & \checkmark &  \\
Teacher Soft Labels                        &\checkmark &   & \checkmark & & \checkmark\\
Mixup                      & \checkmark & \checkmark &  & &   \\
ACM Loss                  & \checkmark &\checkmark  & \checkmark & \checkmark &   \\
\hline
Accuracy              & 90.76 & 92.50 & 91.17 & 92.87 & 89.38 \\
Robustness            & 56.65 & 52.29 & 49.53 & 48.38 & 0.21\\
\hline
\end{tabular}
\end{table}

\subsubsection{Role of Intermediate Features}
To understand the role of low, mid, and high-level features, we performed experiments on CIFAR-10 by progressively changing blocks used for distillation. For this ablation study, we kept all the standard settings reported in the Section A.1. Our correspondence of blocks and features is as follows: block 2: low-level features; block 3: mid-level features; block 4: high-level features. Please note that block 1 corresponds to the output of the first layer only. Therefore, we do not call it low-level features. The results are shown in the Table~\ref{tab:mid_features}.

As shown in the Table~\ref{tab:mid_features}, mid-level features play a more important role in robustness transfer and high-level features play a crucial role in accuracy transfer. The robustness of the student is 39.37\% when we only use mid-level features, but it decreases to 33.13\% when high-level features are utilized alone. On the other hand, clean accuracy improves when we only use high-level features: 88.30\% with mid-level compared with 91.18\% with high-level features. In addition, a combination of mid and high-level features is enough to get close to optimal robustness and accuracy. But the addition of low-level features improves robustness even further.

Apart from these low, mid, and high-level features, we also used the output of the first layer in the proposed loss function. Our experiments above show that the improvement brought by first layer distillation is relatively small. Specifically, the addition of the first layer in the above-mentioned experiments brings $\leq 1$\% improvement for robustness.

In summary, all level features (low, mid, high level) improve robustness and accuracy. However, mid-level features seem to be more critical for robustness and high-level features for accuracy.

\begin{table}
    \centering
    \begin{adjustbox}{width=0.5\linewidth}
    \begin{tabular}{l|cc}
         \hline
         \textbf{Features Used}& \textbf{Accuracy} & \textbf{Robustness} \\
      
         \hline
          Low-Level (2)    & 86.36 & 17.24 \\
          Mid-Level (3)    & 88.30  & 39.37 \\
          High-Level (4)   & 91.18 & 33.13 \\
          \hline
          Low+Mid Level (2+3)      & 88.15 & 41.92 \\
          Mid+High Level (3+4)     & 90.69 & 52.79 \\
          
         \hline
         First Layer Only (1)       & 86.00    & 0.31  \\
         No First Layer (2+3+4)    & 90.69 & 56.15 \\
         \hline
         All Features   & 90.76  & 56.65     \\
         \hline
    \end{tabular}
    \end{adjustbox}
    \caption{Impact of of using different intermediate features on clean accuracy and robustness of student.}
    \label{tab:mid_features}
\end{table}

\subsubsection{Comparison of Losses}
The purpose of ACM loss is to match activated channel maps of teacher and student. It is possible to distill robustness by directly matching intermediate features of teacher and student. However, this direct way of distillation overlooks differences between the teacher and the student such as structure, number of channels, size of activations, how and on what data teacher is trained, etc.

To see the effect of directly distilling the intermediate features, we also have conducted an ablation study comparing direct distillation ($\ell_2$-loss) with ACM-based distillation while progressively increasing differences between the teacher and the student. We have kept all the standard settings (Section A.1) and used similar settings for direct distillation for a fair comparison. 

The results are reported in Table~\ref{tab:losses}. When teacher and student are similar, ACM performs slightly better than direct distillation (56.65\% vs. 56.12\%). However, when the number of channels of teacher and student is different, the performance gap increases (48.75\% vs. 44.95\%). This gap increases further when both channels and the number of layers are different (47.18\% vs. 41.90\%). A similar gap is also visible in terms of clean accuracy for all these cases.

To further explore the effect of this difference, we also performed one experiment under transfer learning settings for CIFAR-100 (Table 5 in the paper). Here, the teacher is trained on a different dataset (ImageNet), so the difference between the two models is larger. The performance gap is also wider. ACM outperforms direct distillation significantly (clean accuracy: 65.69\% vs. 57.86\% and robustness: 24.14\% vs. 16.20\%).

\begin{table}
\centering
\begin{tabular}{lcccc}
\toprule
 Method & Teacher  & Student  & Accuracy & Robustness  \\
 \toprule
\multicolumn{5}{c}{Distillation for CIFAR-10}	\\
\midrule
  Full      & WRN28-10 & WRN34-10 & 89.98  & 56.12\\
  MixACM       & WRN28-10 & WRN34-10  & \textbf{90.76} & \textbf{56.65} \\
\midrule
  Full      & WRN28-10 & WRN28-5 & 88.46 & 44.95 \\
  MixACM       & WRN28-10 & WRN28-5 & \textbf{90.26}   & \textbf{48.75
  }\\
  \midrule
  Full      & WRN-34-20 & WRN16-10   & 84.27 & 41.9 \\
  MixACM.   & WRN-34-20 & WRN16-10   & \textbf{86.31} & \textbf{47.18} \\
\midrule
\multicolumn{5}{c}{Transfer Learning for CIFAR-100}	\\
\midrule
Direct &	WRN28-10&	WRN34-10&	57.86&	16.20\\
ACM	&WRN28-10&	WRN34-10&	65.69	&24.14\\
\bottomrule
\end{tabular}
\caption{Effect of using direct loss vs. MixACM loss on distillation and transfer learning.}
\label{tab:losses}
\end{table}

\subsubsection{Effect of $\alpha_{acm}$} The only extra hyper-parameters introduced by our algorithm is the weight of ACM loss ($\alpha_{acm}$). To see the effect of $\alpha_{acm}$, we perform an ablation experiment with WideResNet-34-10 as student. To avoid any confounding effect of other factors, we use only ACM loss in this experiment. The results are shown in Figure \ref{fig:ablation_study_alpha}.

In summary, the clean accuracy of the model is less sensitive to $\alpha_{acm}$ compared with robustness. For instance when we vary the values of $\alpha_{acm}$ from $100000$ to $100$, the clean accuracy changes from $94$ to $92$ while robustness changes from $50$ to $30$. The value of $\alpha_{acm}$ also acts as a trade-off between clean accuracy and robustness. 

\subsubsection{Effect of $\alpha_{kld}$ and temperature $\gamma$} 
\begin{figure}[t]
\begin{subfigure}[b]{1\textwidth}
    \centering
    \includegraphics[width=1\textwidth]{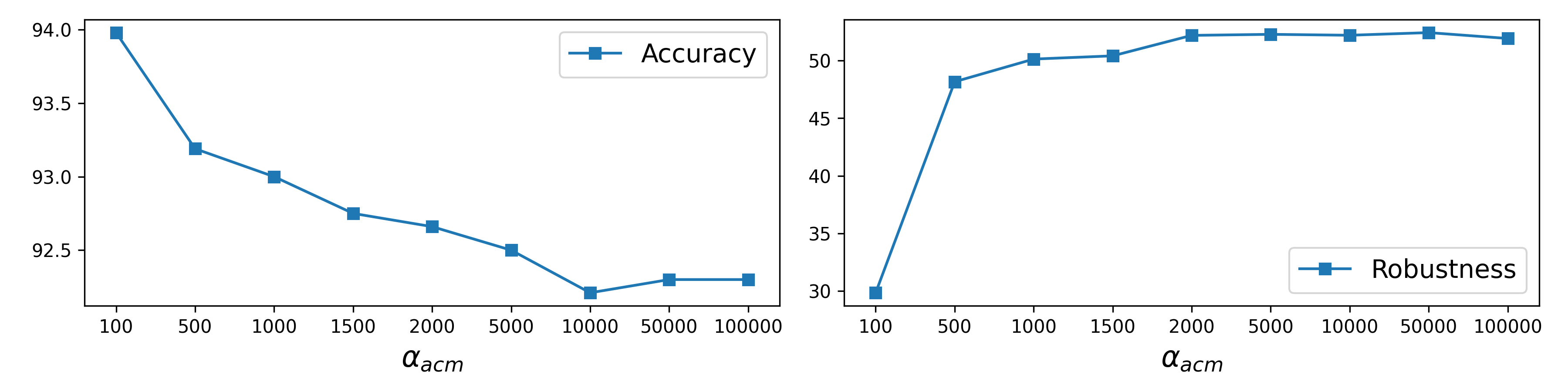}
    \caption{Effect of $\alpha_{acm}$ on accuracy and robustness.}
    \label{fig:ablation_study_alpha}
\end{subfigure}
\hfill
\begin{subfigure}[b]{1\textwidth}
    \centering
    \includegraphics[width=1\textwidth]{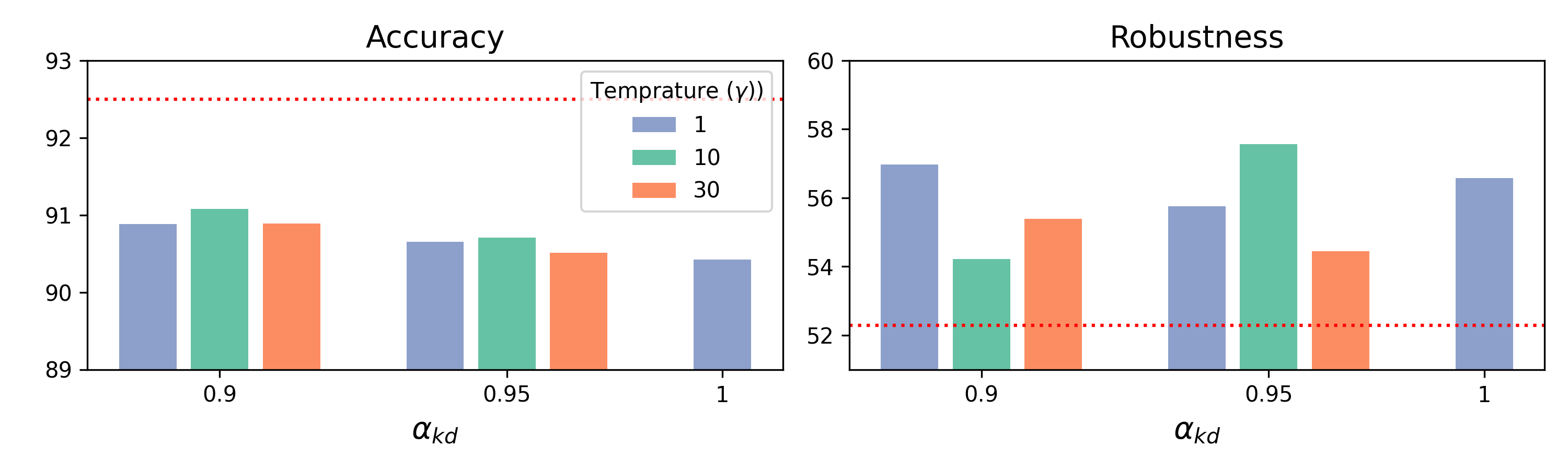}
    \caption{Effect of $\alpha_{kld}$ and temperature $\gamma$ on accuracy and robustness. The dotted red line represents training without soft labels (i.e., $\alpha_{kld}=0$).}
    \label{fig:ablation_study_alpha_kd}
\end{subfigure}
\caption{\textbf{Hyper-parameter Selection.} Comparison of three different hyper-parameters on accuracy and robustness.}

\end{figure}

Our proposed method uses soft labels with KLD loss \cite{hinton2015distilling}. This loss has two hyper-parameters: $\alpha_{kld}$ and temperature $\gamma$. We select best values of these hyper-parameters with an experiment where we set $\alpha_{acm}=5000$ and tried three different settings of these two parameters. The results are reported in Figure~\ref{fig:ablation_study_alpha_kd}. The student is a WideResNet-34-10. We select best values: $\alpha_{kld}=0.95, \gamma=10$. We use these hyper-parameter values for all of our experiments. 

\subsubsection{Comparison of Transforms} Our method requires a transformation function applied on Activated Channel Maps if teacher and student have different number of channels. For this purpose, we compare performance of three different transformation functions: a fully connected layer, an adaptive average pool function or adaptive max pool function. We applied these functions under two settings: applied on teacher's activated channel maps or applied on student's activated channel maps. The results are shown in Table \ref{tab:ablation_transform}.

\begin{table}
\centering
\caption{Effect of different transformation functions for robustness transfer from robust WideResNet-28-10 to WideResNet-28-5.}
\label{tab:ablation_transform}
\begin{tabular}[width=1\textwidth]{l|c|c|c}
\hline
\textbf{Teacher Transform} & \textbf{Student Transform} & \textbf{Accuracy} & \textbf{Robustness} \\
\hline
None & Adaptive Max Pool & 89.92  & 46.41  \\
None & Adaptive Avg Pool & 90.04  & 46.27 \\
None & Affine            & 88.86  & 30.27 \\
\hline
Adaptive Max Pool & None & 90.26  & 48.75 \\
Adaptive Avg Pool & None & 89.99  & 42.01 \\
Affine            & None & 88.39 & 23.51 \\
\hline
\end{tabular}
\end{table}

\begin{table}
    \centering
    \caption{\textbf{Distillation with Different Size Ratios.} The table shows result of distillation from a large teacher to a progressively small student. Size ratio is ratio of number of trainable parameters in student to teacher. Our method works well to reduce teacher size 10$\times$. However, significant compression of depth and width causes a sharp deterioration of performance. }
    \label{tab:distillation_supp}
    \begin{tabular}{llccccc}
    \toprule
         \multirow{2}{*}{Teacher} & \multirow{2}{*}{Student} & \multirow{2}{*}{Size Ratio} & \multicolumn{2}{c}{PGD7-AT} & \multicolumn{2}{c}{Ours} \\
         \cmidrule{4-7}
                        &                &       & Acc.  & Rob.  & Acc.  & Rob. \\
        \midrule
        WRN 28-10  &  WRN 28-10  & 100\%    & 84.17 & 49.23 & 90.48 & 54.05   \\
                   &  WRN 28-5   & 25.04\%  & 83.90 & 47.74 & 90.26 & 48.75   \\
                   &  WRN 28-1   & 1.01\%        & 78.89 & 48.27 & 83.86 & 8.67    \\
                   &  WRN 22-10  & 73.5\%        & 84.26 &  49.29  & 90.40 & 51.23   \\
                   &  WRN 22-5   & 18.40\%        & 83.73 & 48.93 & 89.49 & 46.16   \\
                   &  WRN 22-1   & 0.75\%        & 78.36 & 47.57 & 82.74 & 5.81    \\
        \midrule
                   &  WRN 16-10  & 46.92\%  & 83.90 & 47.55 & 89.93 & 48.09 \\
                   &  WRN 16-5   & 11.76\%  & 83.51 & 48.11 & 88.96 & 36.95 \\
                   &  WRN 16-1   & 0.48\%        & 75.69 & 44.46 & 79.90 & 2.45     \\
                   &  WRN 10-10  & 20.38\%        & 78.59 & 44.41 & 86.31 & 6.54   \\
                   &  WRN 10-5   & 5.12\%        & 78.16 & 44.04 & 82.82 & 3.41    \\
                   &  WRN 10-1   & 0.21\%        & 67.11 & 38.85 & 66.65 & 1.15 \\
    \bottomrule
    \end{tabular}
\end{table}

\section{Theoretical Results}
This appendix has full proofs for results in Section 4. The order is as follows. We first prove that the population adversarial error can be bounded above by the sum of an empirical adversarial loss and a distillation loss. Then we consider a binary classification task with the logistic loss function and prove that the distillation loss can be approximated by the mixup augmentation based loss.

\subsection{Proof of Theorem~4.1}

{\bf Theorem.} {\it We consider task of mapping input $x \in \mathcal{X} \subseteq \mathbb{R}^d$ to label $y \in \mathcal{Y}=\{1,2,\ldots,K\}$. Denote the training data as
$
\cD = \{ (x_1, y_1), ..., (x_n, y_n) \}
$,
where each data point is sampled from a ground truth distribution $P.$
The goal is to learn a classifier $f: \mathcal{X} \to \R^K$ from a hypothetical space $\cF.$ 
Let $L(f(x), y) = 1 - \phi_\gamma\big( f(x), y\big)$ be the loss function, where $\gamma>0$ is the temperature and
\benrr
\phi_\gamma\big( f(x), y\big) = \frac{\exp(f(x)_y / \gamma)}{\sum_{k=1}^K \exp(f(x)_k / \gamma)}.
\eenrr
Suppose a classifier $f$ is decomposed into $g\circ h$, where $h \in \cH$ is a feature extractor and $g \in \cG$ stands for the top model.
We considered supervision of a teacher feature extractor $h^T(x)$ trained on same or similar dataset.
Then, with probability $1-\delta$, for any $f \in \cF$, 
\benr
\E\big[\max_{\|\delta\|_\infty \leq \epsilon} \bbI\{\hat y(x+\delta)\neq y\}\big]
\leq 2\tilde \cL(f, \cD) + \frac{8}{\gamma}\tilde \cL_{dis}(f, \cD)  + 4 \mathfrak{R}_S(\Psi_T) + \frac{16}{n}+ 6\sqrt{\frac{2/\delta}{2n}},
\eenr
where $\mathfrak{R}_S$ is the Rademacher complexity, $\Psi_T = \big\{ \max_{\|\delta\|_\infty<\epsilon} L(g \circ h^T(x+\delta), y): \,\, g\in \cG\big\}$ and
\benrr
\tilde \cL_{dis}(f,\cD) = \frac{1}{n} \inf_{g \in \cG } \sum_{i=1}^n \max_{\|\delta\|_\infty<\epsilon} \big| f(x_i+\delta) - g \circ h^T(x_i+\delta) \big|.
\eenrr
}

{\bf Proof:} 
Denote the adversarial loss with $\epsilon$-bounded $\ell_\infty$ adversarial attacks as
\benrr
\tilde L(f(x), y ) = \max_{\|\delta\|_\infty<\epsilon} L\big(f(x+\delta), y \big).
\eenrr
Let $j \in \{1, \ldots, \log_2(n)\}$ and $\tau_j = 2^{2-j}.$ Here, for simplicity, we assume $\log_2(n)$ is a positive integer. Then, we denote the following classes of functions
$\Psi_T = \{ \tilde L(g \circ h^T(x), y): \,\, g\in \cG\}$ and 

\benrr
\Psi_j = \{\tilde L (f(x), y): \,\, \inf_{g \in \cG} \frac{1}{n}\sum_{i=1}^n |\tilde  L(f(x_i), y_i) - \tilde L (g \circ h^T(x_i), y_i)| \leq \tau_j \}.
\eenrr 
By the classical generalization bound with the Rademacher complexity, with probability at least $1-\delta$, for any $\tilde L \in \Psi_j$,
\benr\label{eq11}
\E\big[ \tilde L ( f(x), y) \big] & \leq & \frac{1}{n} \sum_{i=1}^n \tilde L ( f(x_i), y_i)  + 2 \mathfrak{R}_S(\Psi_j) + 3\sqrt{\frac{2/\delta}{2n}},
\eenr
where
\benrr
\mathfrak{R}_S(\Psi_j) &=& \frac{1}{n} \E_{\vep}\Big[ \sup_{\tilde L (f) \in \Psi_j} \sum_{i=1}^n \vep_i \tilde L \big( f(x_i), y_i \big) \Big].
\eenrr
Furthermore, we have
\benrr
\mathfrak{R}_S(\Psi_j) &=& \frac{1}{n} \E_{\vep}\Big[ \sup_{\tilde L \in \Psi_j} \inf_{g \in \cG} \sum_{i=1}^n \vep_i \tilde L \big( f(x_i), y_i \big) \Big] \\
&=& \frac{1}{n} \E_{\vep}\Big[ \sup_{\tilde L(f) \in \Psi_j} \inf_{g \in \cG} \sum_{i=1}^n \vep_i \big( \tilde L ( f(x_i), y_i) -\tilde L ( g\circ h^T(x_i), y_i) + \tilde L ( g \circ h^T(x_i), y_i) \big) \Big] \\
&\leq& \frac{1}{n} \E_{\vep} \Big[ \sup_{\tilde L(f) \in \Psi_j} \inf_{g \in \cG} \max_i | \vep_i| \sum_{i=1}^n \big| \tilde L ( f(x_i), y_i) -\tilde L ( g \circ h^T(x_i), y_i) \big| \Big] \\
&& + \frac{1}{n} \E_{\vep}\Big[ \sup_{g \in \cG} \sum_{i=1}^n \vep_i \tilde L (g\circ h^T(x_i), y_i)\Big] \\
&\leq & \tau_j + \mathfrak{R}_S(\Psi_T).
\eenrr 
Plugging the upper bound of $\mathfrak{R}_S(\Psi_j)$ into (\ref{eq11}), we have with probability at least $1-\delta$, 
\benr\label{eq12}
\E\big[\tilde L ( f(x), y) \big] & \leq & \frac{1}{n} \sum_{i=1}^n \tilde L ( f(x_i), y_i)  + 2 \tau_j + 2\mathfrak{R}_S(\Psi_T) + 3\sqrt{\frac{2/\delta}{2n}}.
\eenr
Then the inequality (\ref{eq12}) holds for all $f\in\cF$ and $j$ with probability at least $1-\log_2(n) \delta.$
Given $f$, then for any $g \in \cG$, 
\benrr
\frac{1}{n} \sum_{i=1}^n \big| \tilde L ( f(x_i), y_i) -\tilde L ( g \circ h^T(x_i), y_i) \big| \leq 2.
\eenrr
This implies that for any given $f$, there exists $j \in \{1, \ldots, \log_2(n)\}$ such that $\tilde L ( f(x), y) \in \Psi_j.$
We select the smallest $j$ that satisfies
\benrr
\tau_j \geq \frac{1}{n}\inf_{g \in \cG } \sum_{i=1}^n \big| \tilde L ( f(x_i), y_i) - \tilde L ( g \circ h^T(x_i), y_i) \big|  \geq \frac{1}{2} \tau_j - 2^{1-\log_2(n)}.
\eenrr
Then 
\benrr
\tau_j \leq \frac{4}{n} + \frac{2}{n} \inf_{g \in \cG } \sum_{i=1}^n \big| \tilde L ( f(x_i), y_i) -\tilde L ( g \circ h^T(x_i), y_i) \big|.
\eenrr
Furthermore, the inequality (\ref{eq12}) can be rewritten as
\benr\label{res1}
\E\big[ \tilde L ( f(x), y) \big] & \leq & \frac{1}{n} \sum_{i=1}^n \tilde L ( f(x_i), y_i)  + \frac{8}{n}  \nonumber \\
&& + \frac{4}{n} \inf_{g \in \cG } \sum_{i=1}^n \big| \tilde L ( f(x_i), y_i) -\tilde L ( g \circ h^T(x_i), y_i) \big| \nonumber \\
&& + 2 \mathfrak{R}_S(\Psi_T) + 3\sqrt{\frac{2/\delta}{2n}},
\eenr
with probability at least $1-\delta.$
In addition,
\benrr
\tilde L ( f(x), y)  &\leq& \max_{\|\delta\|_\infty<\epsilon} \big| L( f(x), y) -  L( g \circ h^T(x), y) \big| + \tilde L ( g \circ h^T(x), y), \\
\tilde L ( g \circ h^T (x), y)  &\leq& \max_{\|\delta\|_\infty<\epsilon} \big| L( f(x), y) - L( g \circ h^T(x), y) \big| + \tilde L ( f(x), y).
\eenrr
So we have
\benrr
\E\big[ \tilde L ( f(x), y) \big] & \leq & \frac{1}{n} \sum_{i=1}^n \tilde L ( f(x_i), y_i)  + \frac{8}{n}  \nonumber \\
&& + \frac{4}{n} \inf_{g \in \cG } \sum_{i=1}^n \max_{\|\delta\|_\infty<\epsilon}  \big|  L( f(x_i+\delta), y_i) - L( g \circ h^T(x_i+\delta), y_i) \big| \nonumber \\
&& + 2 \mathfrak{R}_S(\Psi_T) + 3\sqrt{\frac{2/\delta}{2n}},
\eenrr

Next we show the relationship between the adversarial accuracy and $\E\big[ \tilde L ( f(x), y) \big].$
The adversarial loss function $\tilde L $ can be rewritten as
\benr\label{losstrans}
\tilde L  (f(x), y) &=& \max_{\|\delta\|_\infty<\epsilon} \frac{\sum_{k \neq y}\exp\big(f(x+\delta)_k/\gamma\big)}{\sum_{k \in [K] } \exp\big(f(x+\delta)_k/\gamma\big)} \nonumber \\
&=& \max_{\|\delta\|_\infty<\epsilon} \frac{1}{1 + \frac{\exp\big(f(x+\delta)_y/\gamma\big)}{\sum_{k \neq y}\exp\big(f(x+\delta)_k/\gamma\big)}} \nonumber \\  
&=& \max_{\|\delta\|_\infty<\epsilon} \frac{1}{1+\exp\big(f(x+\delta)_y/\gamma - \ln(\sum_{k\neq y}\exp(f(x+\delta)_k/\gamma))   \big)} \nonumber \\
&=& \max_{\|\delta\|_\infty<\epsilon} \sigma\Big(-\frac{f(x+\delta)_y}{\gamma} + \ln\big(\sum_{k\neq y} \exp(\frac{f(x+\delta)_k}{\gamma})\big)  \Big),
\eenr
where $\sigma$ is the sigmoid function. Notice that $\sigma$ is a monotonically increasing function and
\benrr
 \ln\big(\sum_{k\neq y} \exp(\frac{f(x+\delta)_k}{\gamma})\big) \geq \max_{k \neq y} \frac{f(x+\delta)_k}{\gamma}.
\eenrr
Then we have
\benr
\tilde L  (f(x), y) \geq  \max_{\|\delta\|_\infty<\epsilon} \sigma\Big(-\frac{f(x+\delta)_y}{\gamma} + \max_{k \neq y} \frac{f(x+\delta)_k}{\gamma} \Big)
\eenr
If there exists $\delta$ such that $f(x+\delta)_y \leq \max_{k \neq y} f(x+\delta)_k$, then
\benr\label{eq13}
 \max_{\|\delta\|_\infty<\epsilon} \sigma\Big(-\frac{f(x+\delta)_y}{\gamma} + \max_{k \neq y} \frac{f(x+\delta)_k}{\gamma} \Big) \geq \frac{1}{2}  \max_{\|\delta\|_\infty<\epsilon} \bbI\big( f(x+\delta)_y \leq \max_{k \neq y} f(x+\delta)_k \big).
\eenr
In contrast, if for any $\delta$, $f(x+\delta)_y > \max_{k \neq y} f(x+\delta)_k$, then 
\benr\label{eq14}
\sigma\Big(-\frac{f(x+\delta)_y}{\gamma} + \max_{k \neq y} \frac{f(x+\delta)_k}{\gamma} \Big) \geq \bbI\big( f(x+\delta)_y \leq \max_{k \neq y} f(x+\delta)_k \big).
\eenr
Combining (\ref{eq13}) and (\ref{eq14}),
\benr\label{res2}
2 \tilde L  (f(x), y) \geq  \max_{\|\delta\|_\infty<\epsilon} \bbI\big( f(x+\delta)_y \leq \max_{k \neq y} f(x+\delta)_k \big).
\eenr
According to (\ref{res1}) and (\ref{res2}), we have
\benrr
&& \sP\big[ \{(x, y): \exists \delta, \,\, \|\delta\|_\infty < \epsilon \,\, \text{s.t.} \,\, \hat y(x+\delta)\neq y \} \big] \\
&=&\E\big[ \max_{\|\delta\|_\infty<\epsilon}  \bbI\big( f(x+\delta)_y \leq \max_{k \neq y} f(x+\delta)_k \big) \big] \\
&\leq&  2 \E\big[ \tilde L  (f(x), y) \big] \\
&\leq & \frac{2}{n} \sum_{i=1}^n \tilde L ( f(x_i), y_i)  + \frac{16}{n}  \nonumber \\
&& + \frac{8}{n} \inf_{g \in \cG } \sum_{i=1}^n \max_{\|\delta\|_\infty<\epsilon}  \big|  L( f(x_i+\delta), y_i) - L( g \circ h^T(x_i+\delta), y_i) \big| \nonumber \\
&& + 4 \mathfrak{R}_S(\Psi_T) + 6\sqrt{\frac{2/\delta}{2n}},
\eenrr
with probability at least $1-\delta.$ According to \cite{gao2017properties}, 
\benrr
&& \sP\big[ \{(x, y): \exists \delta, \,\, \|\delta\|_\infty < \epsilon \,\, \text{s.t.} \,\, \hat y(x+\delta)\neq y \} \big] \\
&\leq & \frac{2}{n} \sum_{i=1}^n \tilde L ( f(x_i), y_i)  + \frac{16}{n}  \nonumber \\
&& + \frac{8}{n \gamma} \inf_{g \in \cG } \sum_{i=1}^n \max_{\|\delta\|_\infty<\epsilon}  \big| f(x_i+\delta) -  g \circ h^T(x_i+\delta) \big| \nonumber \\
&& + 4 \mathfrak{R}_S(\Psi_T) + 6\sqrt{\frac{2/\delta}{2n}}.
\eenrr

\bbox

\subsection{Proof of Theorem~4.2}

In this section, we start with a binary classification task and logistic regression. Denote
\benrr
f_\theta(x) = \theta^\top x, \quad g(f_\theta(x)) = \frac{1}{1+\exp(-f_\theta(x))}, \quad h(f_\theta(x)) = \log\big(1+\exp(f_\theta(x))\big).
\eenrr
Then the loss function $L$ can be rewritten as
\benrr
L(f_\theta(x),y) = h(f_\theta(x)) - y f_\theta(x).
\eenrr
Notice that $\|\delta\|_\infty \leq \epsilon$ implies $\| \delta \|_2 \leq \epsilon \sqrt d$, and therefore
\benrr
\max_{\|\delta\|_\infty \leq \epsilon} L( f_\theta(x), y) \leq 
\max_{\| \delta \|_2 \leq \epsilon \sqrt d}L( f_\theta(x), y),
\eenrr
where $d$ is the dimension of the input $x.$ The standard empirical loss function can be written as 
\benrr
\cL (f_\theta, \cD) = \frac{1}{n} \sum_{i=1}^n L\big(f_\theta(\rvx_i), \rvy_i \big) = \frac{1}{n} \sum_{i=1}^n h\big(f_\theta(\rvx_i)\big) - \rvy_i f_\theta(\rvx_i),
\eenrr
where $\cD = \{(\rvx_i, \rvy_i), i=1,\ldots, n\}.$
For a given $\epsilon>0$, we consider the adversarial loss with $l_2$-attack of size $\epsilon\sqrt{d}$, that is,
\benrr
\tilde \cL(f_\theta, \cD) = \frac{1}{n} \sum_{i=1}^n \max_{\|\delta_i\|_2\leq \epsilon\sqrt{d}} \tilde L \big( f_\theta(\rvx_i+\delta_i), \rvy_i \big)=\frac{1}{n}\sum_{i=1}^n \tilde L \big(f_\theta(\rvx_i),\rvy_i\big)
\eenrr
Consider the data-dependent parameter space:
\benrr
\Theta &=& \{\theta \in \R^d: \rvy_i f_\theta(\rvx_i) + (\rvy_i -1)f_\theta(\rvx_i) \geq 0, \\
&& \quad \text{and} \quad |\rvy_i^* - g(f_\theta(\rvx_i))| \leq \beta |\rvy_i - g(f_\theta(\rvx_i))|, \,\, \text{for all}\,\, i = 1, \ldots, n \}.
\eenrr
The first inequality considers the zero training error ($0$-$1$ loss). That is 
$f_\theta(\rvx)>0$ when $\rvy =1$ and $f_\theta(\rvx)\leq 0$ if $\rvy =0.$ The second inequality constraints the distillation, i.e. $g(f_\theta(\rvx_i)$ is closed to the soft label given by the teacher model. 
Now we are ready to state the following theorem:

{\bf Theorem.} {\it Suppose there exists a constant $c_x > 0$ such that $\|\rvx_i\|_2 > c_x \sqrt d$ for all $i \in \{1, \cdots, n \}.$ 
Then, for any $\theta \in \Theta$, we have
\benrr
\tilde \cL(f_\theta, \cD)  + \alpha \tilde \cL(f_\theta, \cD_{dis}) \leq  \cL_{mix}(f_\theta, \cD) + \alpha \cL_{mix}(f_\theta, \cD_{dis}),
\eenrr
where the size of the adversarial attack $\epsilon$ is 
\benrr
\epsilon = \frac{1-\alpha \beta}{1+\beta} c_x R\,\, \E_{\lambda\sim \tilde P_{\lambda}}[1-\lambda], \quad \text{with} \quad R = \min_{i\in\{1,\ldots,n\}} |\cos(\theta, \rvx_i)|,
\eenrr
and the distribution $\tilde P_{\lambda}$ is
\benrr
\tilde P_{\lambda} (\lambda) = \frac{\alpha}{\alpha+\beta}\text{Beta}(\alpha+1,\beta) + \frac{\beta}{\alpha+\beta} \text{Beta}(\beta+1, \alpha).
\eenrr
}

{\bf Proof:} By the second order Taylor approximation,
\benrr
L\big(f_\theta (x+\delta), y \big) \approx L\big( f_\theta(x), y\big) + \delta^\top \frac{\pa}{\pa x} L\big(f_\theta(x), y\big)  + \frac{1}{2} \delta^\top \frac{\pa^2}{\pa x \pa x^\top} L\big(f_\theta(x), y\big) \delta.
\eenrr
Note that
\benrr
\frac{\pa}{\pa x} L\big(f_\theta(x), y\big) &=& \frac{\pa}{\pa x}  \big( h(f_\theta(x)) - y f_\theta(x) \big) \\
&=& \frac{\pa}{\pa f} h(f_\theta(x)) \frac{\pa}{\pa x} f_\theta(x) - y \frac{\pa}{\pa x} f_\theta(x)\\
&=& g\big(f_\theta(x)\big) \theta - y\theta \\
&=& \big(g(\theta^\top x) - y\big) \theta
\eenrr
and
\benrr
\frac{\pa^2}{\pa x \pa x^\top} L\big(f_\theta(x), y\big) &=& \frac{\pa^2}{\pa x \pa x^\top} \big( h(f_\theta(x)) - y f_\theta(x) \big) \\
&=& \frac{\pa^2}{\pa f \pa f} h(f_\theta(x)) (\frac{\pa}{\pa x} f_\theta(x))^2 \\
&=& \frac{\pa}{\pa f} g(f_\theta(x)) \theta \theta^\top \\
&=& g(\theta^\top x)\big(1-g(\theta^\top x)\big) \theta \theta^\top.
\eenrr
So we have
\benrr
L\big(f_\theta(x+\delta), y\big) \approx L\big(f_\theta(x), y\big) + \big(g(\theta^\top x) - y\big) \theta^\top \delta + \frac{1}{2}g(\theta^\top x)\big(1-g(\theta^\top x)\big) (\theta^\top \delta)^2
\eenrr
Furthermore
\benrr
\tilde \cL(f_\theta, \cD) &\approx& \frac{1}{n}\sum_{i=1}^n L\big(f_\theta(\rvx_i), \rvy_i\big) + \frac{1}{n}\sum_{i=1}^n \max_{\|\delta_i\|_2\leq \epsilon\sqrt{d}} \Big\{ \big(g(\theta^\top \rvx_i) - \rvy_i\big) \theta^\top \delta_i \\
&& + \frac{1}{2} g(\theta^\top \rvx_i)\big(1-g(\theta^\top \rvx_i)\big) (\theta^\top \delta_i)^2\Big\} \\
&=:& \frac{1}{n}\sum_{i=1}^n L\big(f_\theta(\rvx_i), \rvy_i\big) + I_1.
\eenrr
Similarly, for the data $\cD_{dis} = \{(\rvx_i, \rvy_i^*), \rvy^*_i = f^T(\rvx_i), i=1,\ldots, n\}$,
\benrr
\tilde \cL(f_\theta, \cD_{dis}) &\approx& \frac{1}{n}\sum_{i=1}^n L\big(f_\theta(\rvx_i), \rvy_i^*\big) + \frac{1}{n}\sum_{i=1}^n \max_{\|\delta_i\|_2\leq \epsilon\sqrt{d}} \Big\{ \big(g(\theta^\top \rvx_i) - \rvy_i^*\big) \theta^\top \delta_i \\
&& + \frac{1}{2} g(\theta^\top \rvx_i)\big(1-g(\theta^\top \rvx_i)\big) (\theta^\top \delta_i)^2\Big\} \\
&=:& \frac{1}{n}\sum_{i=1}^n L\big(f_\theta(\rvx_i), \rvy_i^*\big) + I_2.
\eenrr
Therefore,
\benrr
\tilde \cL(f_\theta, \cD)  + \alpha \tilde \cL(f_\theta, \cD_{dis}) = \frac{1}{n}\sum_{i=1}^n L\big(f_\theta(\rvx_i), \rvy_i\big) + \frac{\alpha}{n}\sum_{i=1}^n L\big(f_\theta(\rvx_i), \rvy_i^*\big) + I_1 + \alpha I_2. 
\eenrr
Furthermore
\benrr
I_1 + \alpha I_2 &\leq&  \frac{1}{n}\sum_{i=1}^n \max_{\|\delta_i\|_2\leq \epsilon\sqrt{d}} \Big( \big(g(\theta^\top \rvx_i) - \rvy_i\big) \theta^\top \delta_i \Big) \\
&& + \frac{\alpha}{n}\sum_{i=1}^n \max_{\|\delta_i\|_2\leq \epsilon\sqrt{d}} \Big( \big(g(\theta^\top \rvx_i) - \rvy_i^*\big) \theta^\top \delta_i \Big) \\
&& + (1+ \alpha) \frac{1}{2n}\sum_{i=1}^n \max_{\|\delta_i\|_2\leq \epsilon\sqrt{d}} g(\theta^\top \rvx_i)\big(1-g(\theta^\top \rvx_i)\big) (\theta^\top \delta_i)^2 \\ 
&=&  \frac{\epsilon \sqrt d}{n}\sum_{i=1}^n \big| g(\theta^\top \rvx_i) - \rvy_i\big| \|\theta\|_2 + \frac{\alpha \epsilon \sqrt d}{n}\sum_{i=1}^n  \big|g(\theta^\top \rvx_i) - \rvy_i^*\big| \|\theta\|_2 \\
&& + (1+ \alpha) \epsilon^2 d \frac{1}{2n}\sum_{i=1}^n g(\theta^\top \rvx_i)\big(1-g(\theta^\top \rvx_i)\big) \|\theta \|_2^2.
\eenrr

Next we shall bound $\tilde \cL(f_\theta, \cD)  + \alpha \tilde \cL(f_\theta, \cD_{dis})$ by mixup augmentation. Consider the following loss:
\benrr
\cL_{mix}(f_\theta, \cD) &=& \frac{1}{n^2} \sum_{i=1}^n \sum_{j=1}^n \E_{\lambda\sim P_\lambda} \big[ L(f_\theta(\rvx_{ij}(\lambda)), \rvy_{ij}(\lambda)) \big], \\
\cL_{mix}(f_\theta, \cD_{dis}) &=& \frac{1}{n^2} \sum_{i=1}^n \sum_{j=1}^n \E_{\lambda\sim P_\lambda} \big[ L(f_\theta(\rvx_{ij}(\lambda)), \rvy_{ij}^*(\lambda)) \big], \\
\eenrr
where $P_\lambda$ is a Beta distribution $\text{Beta}(\alpha, \beta)$, $\rvy_{ij}(\lambda) = \lambda \rvy_i + (1-\lambda) \rvy_j$ and $\rvy_{ij}^*(\lambda) = \lambda \rvy^*_i + (1-\lambda) \rvy^*_j.$ 
We start with $\cL_{mix}(f_\theta, \cD)$:
\benrr
\cL_{mix}(f_\theta, \cD) &=& \frac{1}{n^2} \sum_{i=1}^n \sum_{j=1}^n \E_{\lambda\sim P_\lambda} \big[ h(\theta^\top \rvx_{ij}(\lambda)) - \rvy_{ij}(\lambda) \theta^\top \rvx_{ij}(\lambda) \big] \\
&=& \frac{1}{n^2} \sum_{i=1}^n  \sum_{j=1}^n \E_{\lambda\sim P_\lambda} \Big\{ \lambda \big[ h(\theta^\top \rvx_{ij}(\lambda)) -  \rvy_i \theta^\top \rvx_{ij}(\lambda)\big] \\
&& + (1-\lambda) \big[ h(\theta^\top \rvx_{ij}(\lambda)) -  \rvy_j \theta^\top \rvx_{ij}(\lambda)\big]\Big\}
\eenrr
We introduce a $0$-$1$ random variable $B$ such that the conditional distribution of $B$ given $\lambda$ is
\benrr
P(B=1|\lambda) = \lambda, \quad \text{and} \quad P(B=0|\lambda) = 1-\lambda.
\eenrr
Rewrite $\cL_{mix}(f_\theta, \cD)$ as
\benrr
\cL_{mix}(f_\theta, \cD) &=& \frac{1}{n^2} \sum_{i=1}^n  \sum_{j=1}^n \E_{\lambda\sim P_\lambda} \Big\{\E_{B|\lambda}\Big\{ B \big[ h(\theta^\top \rvx_{ij}(\lambda)) -  \rvy_i \theta^\top \rvx_{ij}(\lambda)\big] \\
&& + (1-B) \big[ h(\theta^\top \rvx_{ij}(\lambda)) -  \rvy_j \theta^\top \rvx_{ij}(\lambda)\big]\Big\}\Big\}.
\eenrr
Notice that
\benrr
&& P(\lambda, B=1) = P(B=1|\lambda) P(\lambda) =\frac{\lambda^\alpha(1-\lambda)^{\beta-1}}{B(\alpha, \beta)} = \frac{\lambda^\alpha(1-\lambda)^{\beta-1}}{B(\alpha+1, \beta)} \times \frac{\alpha}{\alpha+\beta} \\
&& P(\lambda, B=0) = P(B=0|\lambda) P(\lambda) = \frac{\lambda^{\alpha-1} (1-\lambda)^{\beta}}{B(\alpha, \beta)} = \frac{\lambda^{\alpha-1}(1-\lambda)^{\beta}}{B(\alpha, \beta+1)} \times \frac{\beta}{\alpha+\beta}
\eenrr
where $B(\cdot, \cdot)$ is the Beta function. Thus the marginal distribution of $B$ is
\benrr
P(B=1) = \frac{\alpha}{\alpha+\beta} \quad \text{and} \quad P(B=0) = \frac{\beta}{\alpha+\beta}
\eenrr
and the conditional distribution of $\lambda$ given $B$ is
\benrr
P(\lambda|B=1) = \text{Beta}(\alpha+1, \beta) \quad \text{and} \quad P(\lambda|B=0) = \text{Beta}(\alpha, \beta+1). 
\eenrr
Then we have
\benrr
\cL_{mix}(f_\theta, \cD) &=& \frac{1}{n^2} \sum_{i=1}^n  \sum_{j=1}^n \E_{B} \Big\{\E_{\lambda|B}\Big\{ B \big[ h(\theta^\top \rvx_{ij}(\lambda)) -  \rvy_i \theta^\top \rvx_{ij}(\lambda)\big] \\
&& + (1-B) \big[ h(\theta^\top \rvx_{ij}(\lambda)) -  \rvy_j \theta^\top \rvx_{ij}(\lambda)\big]\Big\}\Big\} \\
&=&  \frac{1}{n^2} \sum_{i=1}^n  \sum_{j=1}^n \Big\{ \frac{\alpha}{\alpha+\beta} \E_{\lambda\sim \text{Beta}(\alpha+1,\beta)} \big[ h(\theta^\top \rvx_{ij}(\lambda)) -  \rvy_i \theta^\top \rvx_{ij}(\lambda)\big]\\
&& + \frac{\beta}{\alpha+\beta} \E_{\lambda\sim \text{Beta}(\alpha,\beta+1)} \big[ h(\theta^\top \rvx_{ij}(\lambda)) -  \rvy_j \theta^\top \rvx_{ij}(\lambda)\big] \Big\}.
\eenrr
In addition, let 
\benrr
\lambda' = 1-\lambda \sim \text{Beta}(\beta+1, \alpha). 
\eenrr
So we can transform the random variable from $\lambda$ to $\lambda'$ and have
\benrr
&& \E_{\lambda\sim \text{Beta}(\alpha,\beta+1)} \big[ h(\theta^\top \rvx_{ij}(\lambda)) -  \rvy_j \theta^\top \rvx_{ij}(\lambda)\big] \\
&=& \E_{\lambda'\sim \text{Beta}(\beta+1,\alpha)} \big[ h(\theta^\top \rvx_{ji}(\lambda')) -  \rvy_j \theta^\top \rvx_{ji}(\lambda')\big].
\eenrr
Plug this equation into $\cL_{mix}(f_\theta, \cD)$,
\benrr
\cL_{mix}(f_\theta, \cD) &=& \frac{1}{n^2} \sum_{i=1}^n  \sum_{j=1}^n \Big\{ \frac{\alpha}{\alpha+\beta} \E_{\lambda\sim \text{Beta}(\alpha+1,\beta)} \big[ h(\theta^\top \rvx_{ij}(\lambda)) -  \rvy_i \theta^\top \rvx_{ij}(\lambda)\big]\\
&& + \frac{\beta}{\alpha+\beta} \E_{\lambda\sim \text{Beta}(\beta+1,\alpha)} \big[ h(\theta^\top \rvx_{ij}(\lambda)) -  \rvy_i \theta^\top \rvx_{ij}(\lambda)\big] \Big\} \\
&=&  \frac{1}{n^2} \sum_{i=1}^n  \sum_{j=1}^n \E_{\lambda\sim \tilde P_{\lambda}} \big[ h(\theta^\top \rvx_{ij}(\lambda)) -  \rvy_i \theta^\top \rvx_{ij}(\lambda)\big],
\eenrr
where
\benrr
\tilde P_{\lambda} (\lambda) = \frac{\alpha}{\alpha+\beta}\text{Beta}(\alpha+1,\beta) + \frac{\beta}{\alpha+\beta} \text{Beta}(\beta+1, \alpha).
\eenrr

To proceed further, we denote $P_n(x)$ as the empirical distribution of $\rvx$ induced by the training sample. Then $I_2$ can be rewritten as
\benrr
\cL_{mix}(f_\theta, \cD) &=& \frac{1}{n} \sum_{i=1}^n \E_{\rvx\sim P_n} \E_{\lambda\sim \tilde P_{\lambda}} \Big[ h(\theta^\top (\lambda \rvx_i +(1-\lambda)\rvx)) \\
&& -  \rvy_i \theta^\top (\lambda \rvx_i +(1-\lambda)\rvx)\Big] \\
&=& \frac{1}{n} \sum_{i=1}^n \E_{\rvx\sim P_n} \E_{\lambda\sim \tilde P_{\lambda}} \psi_i(\gamma), 
\eenrr
where $\gamma = 1-\lambda$ and
\benrr
\psi_i(\gamma) = h(\theta^\top ( (1-\gamma) \rvx_i + \gamma \rvx)) - \rvy_i \theta^\top ( (1-\gamma) \rvx_i + \gamma \rvx).
\eenrr
By the second order Taylor expansion, 
\benrr
\psi_i(\gamma) = \psi_i(0) + \psi'_i(0) \gamma + \frac{1}{2} \psi''_i(0) \gamma^2 + O(\gamma^3).
\eenrr
Furthermore,
\benrr
\psi'_i(0) &=& h'(\theta^\top \rvx_i) \theta^\top (\rvx - \rvx_i) -\rvy_i \theta^\top (\rvx - \rvx_i) \\
&=& \big( g(\theta^{\T}\rvx_i) - \rvy_i \big) \theta^\top (\rvx - \rvx_i),
\eenrr
and
\benrr
\psi''_i(0) &=&  h''(\theta^\top \rvx_i) [\theta^\top(\rvx - \rvx_i)]^2 \\
&=& g(\theta^{\T}\rvx_i) \big( 1-g(\theta^{\T}\rvx_i) \big) [\theta^\top(\rvx - \rvx_i)]^2. 
\eenrr
Thus we have
$ \cL_{mix}(f_\theta, \cD) = \frac{1}{n} \sum_{i=1}^n L(f_\theta(\rvx_i), \rvy_i) + I_{3} + I_{4}$, where 
\benrr
I_3 &=& \E_{\lambda\sim \tilde P_{\lambda}}[1-\lambda] \frac{1}{n} \sum_{i=1}^n \E_{\rvx\sim P_n} \big[ \big( g(\theta^{\T}\rvx_i) - \rvy_i  \big) \theta^\top (\rvx - \rvx_i) \big], \\
I_4 &=& \E_{\lambda\sim \tilde P_{\lambda}}[(1-\lambda)^2] \frac{1}{2 n} \sum_{i=1}^n \E_{\rvx\sim P_n} \big[ g(\theta^{\T}\rvx_i) \big( 1-g(\theta^{\T}\rvx_i) \big) [\theta^\top(\rvx - \rvx_i)]^2 \big].\\
\eenrr
Similarly, for the data $\cD_{dis} = \{(\rvx_i, \rvy_i^*), \rvy^*_i = f^T(\rvx_i), i=1,\ldots, n\}$, the following decomposition also holds: $\cL_{mix}(f_\theta, \cD_{dis}) = \frac{1}{n} \sum_{i=1}^n L(f_\theta(\rvx_i), \rvy^{*}_i) + I_5 + I_4$, where
\benrr
I_5 &=& \E_{\lambda\sim \tilde P_{\lambda}}[1-\lambda] \frac{1}{n} \sum_{i=1}^n \E_{\rvx\sim P_n} \big[ \big( g(\theta^{\T}\rvx_i) - \rvy^*_i  \big) \theta^\top (\rvx - \rvx_i) \big].
\eenrr
Therefore, 
\benrr
\cL_{mix}(f_\theta, \cD) + \alpha \cL_{mix}(f_\theta, \cD_{dis}) &=& \frac{1}{n} \sum_{i=1}^n L(f_\theta(\rvx_i), \rvy_i) + \frac{\alpha}{n} \sum_{i=1}^n L(f_\theta(\rvx_i), \rvy^{*}_i) \\
&& + (I_3 + \alpha I_5) + (1+\alpha) I_4.
\eenrr

For the term $I_3 + \alpha I_5 $, we have
\benrr
I_3 + \alpha I_5 &=&  \E_{\lambda\sim \tilde P_{\lambda}}[1-\lambda] \frac{1}{n} \sum_{i=1}^n \big( (1+\alpha) g(\theta^{\T}\rvx_i) - \rvy_i - \alpha \rvy^{*}_i  \big) \theta^\top \big( \E_{\rvx\sim P_n}(\rvx) - \rvx_i \big) \\
&=& \E_{\lambda\sim \tilde P_{\lambda}}[1-\lambda] \frac{1}{n} \sum_{i=1}^n \big(  \rvy_i + \alpha \rvy^{*}_i  - (1+\alpha) g(\theta^{\T}\rvx_i) \big) \theta^\top \rvx_i \\
&=& \E_{\lambda\sim \tilde P_{\lambda}}[1-\lambda] \frac{1}{n} \sum_{i=1}^n \big| \rvy_i + \alpha \rvy^{*}_i  - (1+\alpha) g(\theta^{\T}\rvx_i) \big| \|\theta\|_2 \|\rvx_i\|_2 |\cos(\theta, \rvx_i)|\\
&=& \E_{\lambda\sim \tilde P_{\lambda}}[1-\lambda] \frac{1}{n} \sum_{i=1}^n \big| |\rvy_i - g(\theta^{\T}\rvx_i)| - \alpha |\rvy^{*}_i- g(\theta^{\T}\rvx_i)| \big| \|\theta\|_2 \|\rvx_i\|_2 |\cos(\theta, \rvx_i)|\\
&\geq& R_i c_x \sqrt{d} \,\, \E_{\lambda\sim \tilde P_{\lambda}}[1-\lambda] \frac{1}{n} \sum_{i=1}^n (1-\alpha k) \big|\rvy_i - g(\theta^{\T}\rvx_i)\big| 
\|\theta\|_2 \\
& \geq & \epsilon \sqrt{d} \frac{1}{n}\sum_{i=1}^n (1+k)\big| \rvy_i - g(\theta^{\T}\rvx_i) \big| \|\theta\|_2\\
& \geq & \frac{1}{n}\sum_{i=1}^n \max_{\|\delta_i\|_2\leq \epsilon\sqrt{d}} \Big(\rvy_i - g(\theta^{\T}\rvx_i) \Big)\theta^\top \delta_i + \frac{\alpha}{n}\sum_{i=1}^n \max_{\|\delta_i\|_2\leq \epsilon\sqrt{d}} \Big( \rvy^{*}_i  - g(\theta^{\T}\rvx_i)\Big) \theta^\top \delta_i. 
\eenrr

Now we turn to $I_4$:
\benrr
I_4 &=& \E_{\lambda\sim \tilde P_{\lambda}}[(1-\lambda)^2] \frac{1}{2 n} \sum_{i=1}^n  g(\theta^{\T}\rvx_i) \big( 1-g(\theta^{\T}\rvx_i) \big) \theta^\top \E_{\rvx\sim P_n} [(\rvx - \rvx_i)(\rvx - \rvx_i)^\top] \theta \\
&=& \E_{\lambda\sim \tilde P_{\lambda}}[(1-\lambda)^2] \frac{1}{2 n} \sum_{i=1}^n  g(\theta^{\T}\rvx_i) \big( 1-g(\theta^{\T}\rvx_i) \big) \theta^\top \big(\E_{\rvx\sim P_n}(\rvx\rvx^\top) + \rvx_i\rvx_i^\top \big) \theta\\
&\geq& \E_{\lambda\sim \tilde P_{\lambda}}[(1-\lambda)^2] \frac{1}{2 n} \sum_{i=1}^n  g(\theta^{\T}\rvx_i) \big( 1-g(\theta^{\T}\rvx_i) \big) \theta^\top \rvx_i\rvx_i^\top \theta \\
&=& R_i^2 c_x^2 d \,\,\E_{\lambda\sim \tilde P_{\lambda}}[(1-\lambda)^2] \frac{1}{2 n} \sum_{i=1}^n  g(\theta^{\T}\rvx_i) \big( 1-g(\theta^{\T}\rvx_i) \big) \|\theta\|_2^2 \\
& \geq & \epsilon^2 d \frac{1}{2 n} \sum_{i=1}^n  g(\theta^{\T}\rvx_i) \big( 1-g(\theta^{\T}\rvx_i) \big) \|\theta\|_2^2 \\
& = & \frac{1}{n}\sum_{i=1}^n \max_{\|\delta_i\|_2\leq \epsilon\sqrt{d}} g(\theta^\top \rvx_i)\big(1-g(\theta^\top \rvx_i)\big) (\theta^\top \delta_i)^2.
\eenrr
Combining the results of $I_3 + \alpha I_5$ and $I_4$, we know that
$
I_3 + \alpha I_5 \geq I_1 + \alpha I_2.
$
Furthermore, we have
\benrr
\tilde \cL(f_\theta, \cD)  + \alpha \tilde \cL(f_\theta, \cD_{dis}) \leq  \cL_{mix}(f_\theta, \cD) + \alpha \cL_{mix}(f_\theta, \cD_{dis}). 
\eenrr

\bbox

Next we extend the above theorem to the case of neural networks with ReLU activation and max-pooling. 
We still consider the binary classification task with the logistic loss and take $f_\theta(x)$ to be a fully connected neural network with ReLU activation function or max-pooling:
\benrr
f_\theta(x) =  \rva^\top \sigma \big(W_{N-1}\cdots \sigma(W_2 \sigma(W_1 x))\big),
\eenrr
where $\sigma(\cdot)$ is a nonlinear function that consists of ReLU activation and max pooling, each $W_i$ is a matrix, and $\rva$ is a column  vector: i.e., $\theta$ consists of $\{W_i, i=1,\ldots, W-1\}$ and $\rva.$ 
According to the derivatives of ReLU and max-pooling,
the function $f_\theta$ satisfies that $\nabla^2_x f_\theta(x) = 0$ and $f_\theta(x) = \nabla_x f_\theta(x) ^\top x$ almost everywhere. Therefore a fully connected neural networks with ReLU activation functions and max-pooling can be locally approximated by a linear function. By the results of logistic regression, we have the following theorem:

{\bf Theorem}. 
{\it Suppose $f_\theta(x) = \nabla_x f_\theta(x) ^\top x$, $\nabla^2_x f_\theta(x) = 0$ and there exists a constant $c_x > 0$ such that $\|\rvx_i\|_2 > c_x \sqrt d$ for all $i \in \{1, \cdots, n \}.$ 
Then, for any $\theta \in \Theta$, we have
\benrr
\tilde \cL(f_\theta, \cD)  + \alpha \tilde \cL(f_\theta, \cD_{dis}) \leq  \cL_{mix}(f_\theta, \cD) + \alpha \cL_{mix}(f_\theta, \cD_{dis}),
\eenrr
where the size of the adversarial attack $\epsilon$ is 
\benrr
\epsilon = \frac{1-\alpha \beta}{1+\beta} c_x R\,\, \E_{\lambda\sim \tilde P_{\lambda}}[1-\lambda], \quad \text{with} \quad R = \min_{i\in\{1,\ldots,n\}} |\cos( \nabla_x f_\theta(x), \rvx_i)|,
\eenrr
and the distribution $\tilde P_{\lambda}$ is
\benrr
\tilde P_{\lambda} (\lambda) = \frac{\alpha}{\alpha+\beta}\text{Beta}(\alpha+1,\beta) + \frac{\beta}{\alpha+\beta} \text{Beta}(\beta+1, \alpha).
\eenrr
}

\bbox

\end{document}